\documentclass{article}

\usepackage{iclr2027_conference,times}
\usepackage{amsmath}
\usepackage{amssymb}
\usepackage{booktabs}
\usepackage{colortbl}
\usepackage{graphicx}
\usepackage{capt-of}
\usepackage{float}
\usepackage[ruled,vlined,linesnumbered]{algorithm2e}
\usepackage{tabularx}
\usepackage{array}
\usepackage{url}
\usepackage{arydshln}
\usepackage{multirow}
\usepackage[most]{tcolorbox}
\usepackage{enumitem}
\usepackage{placeins}
\usepackage{wrapfig}
\usepackage{needspace}
\usepackage{xcolor}
\usepackage[colorlinks=true, citecolor=blue]{hyperref}

\newlength{\workforgeHeadheightDelta}
\fancypagestyle{workforgeFirst}{%
  \fancyhead{}%
  \fancyhead[L]{\includegraphics[height=20pt]{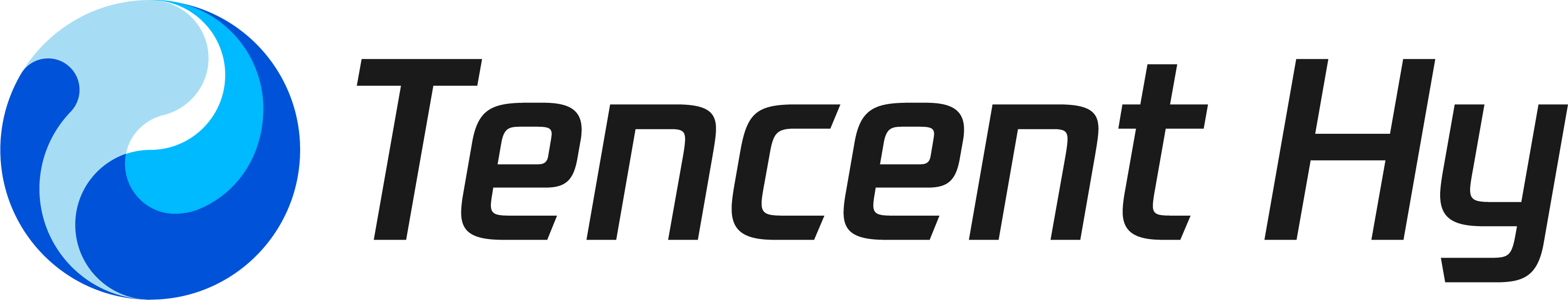}}%
  \fancyhead[R]{\includegraphics[height=30pt]{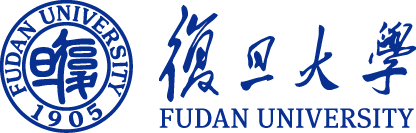}}%
}
\fancypagestyle{workforgeBody}{%
  \fancyhead{}%
}

\newtcblisting{promptbox}[1]{
  enhanced,
  breakable,
  listing only,
  listing engine=listings,
  colback=black!3,
  colframe=black!65,
  colbacktitle=black!80,
  coltitle=white,
  fonttitle=\bfseries,
  title={#1},
  boxrule=0.5pt,
  arc=1mm,
  left=1mm,
  right=1mm,
  top=1mm,
  bottom=1mm,
  before skip=6pt,
  after skip=8pt,
  listing options={
    basicstyle=\ttfamily\scriptsize,
    breaklines=true,
    breakatwhitespace=false,
    columns=fullflexible,
    keepspaces=true,
    showstringspaces=false,
    tabsize=2
  }
}

\graphicspath{{../figure/}{../../../figure/}{../../figure/}{figure/}}

\newcolumntype{Y}{>{\raggedright\arraybackslash}X}
\newlength{\benchmarkcolwidth}
\newlength{\benchmarkcandidatewidth}
\definecolor{sectionGreen}{HTML}{EBEDE8}
\definecolor{statsGroupGreen}{RGB}{208,217,204}
\definecolor{statsGroupPink}{HTML}{F8ECE9}
\definecolor{lightBlue}{RGB}{216,228,245}

\title{Scaling Verifiable Environments for Long-horizon Work Agents}

\author{\textbf{Jiazheng Zhang}{\footnotesize $^{1,2}$}\thanks{Equal Contribution   \ \ $^{\dagger}$ Corresponding Author} 
 \;
 \textbf{Long Ma}{\footnotesize $^{2,*}$}
 \;
 \textbf{Yunxian Yang}{\footnotesize $^{2,*}$}
 \;
 \textbf{Zhiheng Xi}{\footnotesize $^{2, \dagger}$}
 \;
 \\
  \textbf{Zhikai Lei}{\footnotesize $^{2}$}
 \;
 \textbf{Yajie Yang}{\footnotesize $^{2}$}
 \;
 \textbf{Chenyang Liao}{\footnotesize $^{2}$}
 \;
 \textbf{Enyu Zhou}{\footnotesize $^{2}$}
 \;
  \textbf{Yang Nan}{\footnotesize $^{2}$}
 \;
   \textbf{Yuchen Tian}{\footnotesize $^{1}$}
 \;
 \\
 \textbf{Senjie Jin}{\footnotesize $^{2}$}
 \;
 \textbf{Yibo Wang}{\footnotesize $^{1}$}
 \;
 \textbf{Wei He}{\footnotesize $^{2}$}
 \;
 \textbf{Boyang Liu}{\footnotesize $^{2}$}
 \;
 \textbf{Jixuan Huang}{\footnotesize $^{2}$}
 \;
 \textbf{Xin Guo}{\footnotesize $^{2}$}
 \;
 \\
 \textbf{Shutong Jiang}{\footnotesize $^{1}$}
 \;
 \textbf{Zhezheng Hao}{\footnotesize $^{1}$}
 \;
 \textbf{Xinbing Liang}{\footnotesize $^{1}$}
 \;
 \textbf{Zhihao Zhang}{\footnotesize $^{2}$}
 \;
 \textbf{Changzhi Zhou}{\footnotesize $^{1}$}
 \;
 \\
  \textbf{Wiggin Zhou}{\footnotesize $^{1}$}
 \;
 \textbf{Tao Gui}{\footnotesize $^{2,\dagger}$}
 \;
 \textbf{Qi Zhang}{\footnotesize $^{2,\dagger}$}
 \;
\textbf{Xuanjing Huang}{\footnotesize $^2$} 
 \;
\textbf{Clarenceai}{\footnotesize $^{1,\dagger}$} 
 \;
 \;
\\[2pt]
{\footnotesize $^1$} LLM Department, Tencent \;
  {\footnotesize $^2$} Fudan University \; \\
  \texttt{\{jzzhang24, zhxi22\}@m.fudan.edu.cn \ 
 \{tgui, qz\}@fudan.edu.cn}\\
  \texttt{clarenceai@tencent.com}\\
}

\iclrfinalcopy

\begin{document}

\maketitle

\thispagestyle{workforgeFirst}
\pagestyle{workforgeBody}

\begin{abstract}
Work agents operate over digital artifacts to execute professional knowledge-intensive work, requiring training environments that support long-horizon interaction and trustworthy verification.
However, hand-crafted environments incur prohibitive engineering overhead that prevents environment scaling, whereas synthesis methods sacrifice workspace complexity, realism, or grounded verifiability.
To bridge this gap, we introduce \textsc{WorkForge}, a scalable synthesis framework for constructing verifiable work-agent environments from real-world resources.
Starting from expert workflows, \textsc{WorkForge} first identifies the resources, decisions, and deliverables required by each workflow.
It then retrieves relevant real-world files and organizes them into a workspace.
\textsc{WorkForge} inspects the workspace to extract concrete, checkable facts about its content.
These factual anchors fix which task types the workspace can support and how their outcomes can be verified.
Therefore, \textsc{WorkForge} derives each task's instructions, solution plan, and complementary programmatic and semantic verifiers directly from these factual anchors, keeping verification traceable to observable workspace evidence.
Furthermore, we construct 16.7K verifiable environments across 40 professional domains, with workspaces collectively covering 60 file types.
Post-training Qwen3.5-35B-A3B-Base improves GDPval from 45.5 to 73.6 and APEX Score from 5.0 to 21.3, while enabling Qwen3.5-27B to achieve highly competitive performance and outperform strong competitors. 
Our analyses confirm the efficacy of the proposed method and reveal consistent scaling behaviors across both data volume and interaction horizons.

\end{abstract}

\section{Introduction}
\label{sec:introduction}

Autonomous agents mark a fundamental paradigm shift in large language models, transitioning foundation models from passive conversational assistants to active digital workers capable of executing complex, multi-step professional knowledge work \citep{drouin2024workarena,boisvert2024workarenapp,huang2024mlagentbench,patwardhan2026gdpval,vidgen2026apex}.
Agent assistants such as Claude Code and Codex illustrate this paradigm through iterative exploration, tool orchestration, and execution over digital workspaces \citep{anthropic2025claudecode,openai2025codex}.

Unlike conversational or API-calling routines, professional workflows center on concrete deliverables: users specify objectives, and the solution path is embedded within the workspace~\citep{tang2026workspacebench}.
Producing these deliverables requires agents to untangle multiple file dependencies and make decisions in authentic workspace artifacts~\citep{zhou2024webarena,trivedi2024appworld,zhang2026entropy, xi2026agentgymrl}.
Cultivating agents capable of such long-horizon workflow fundamentally demands agent environments that couple authentic operational friction with trustworthy, verifiable supervision \citep{Dou:2025:Sandbox,zhang:2025:CRM,sun2026agentslastexam,xi2026agentgym2}.

While expert-curated environments offer high fidelity, their prohibitive overhead severely impedes environment scaling \citep{zhou2024webarena,trivedi2024appworld}.

\begin{figure}[t]
    \centering
    \includegraphics[width=\linewidth]{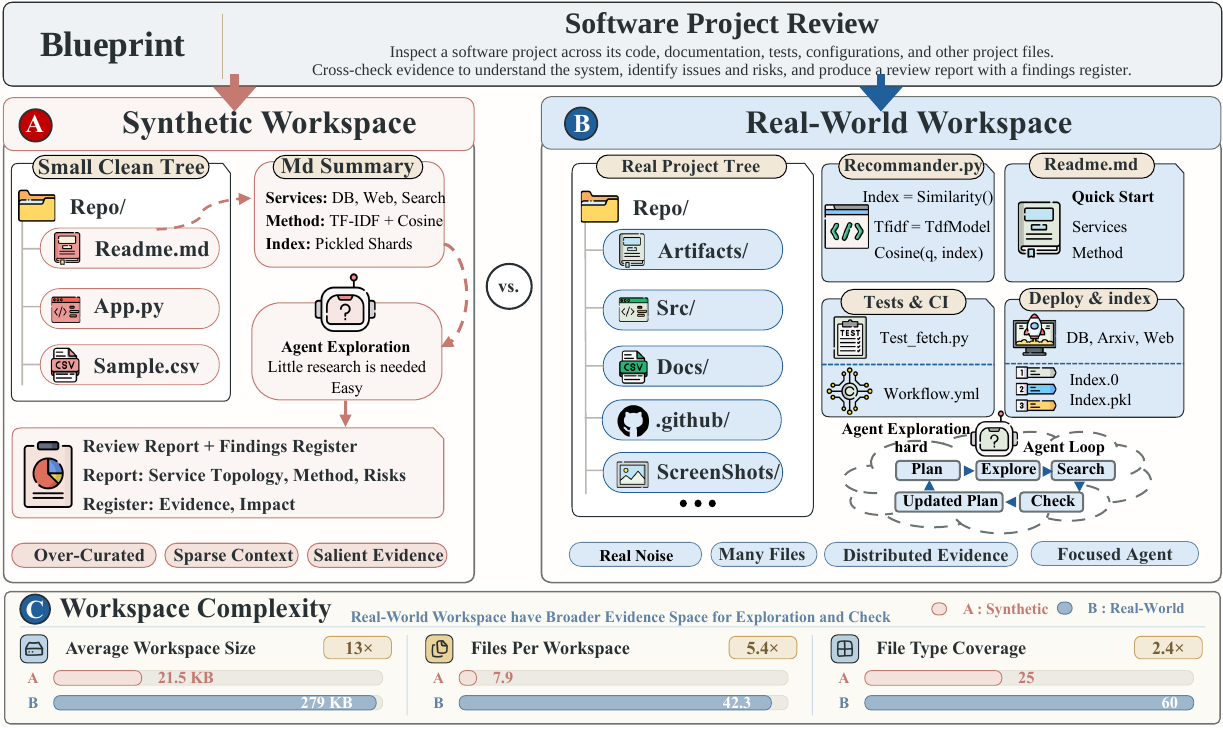}
    \caption{ \textbf{Comparison of synthetic and realistic workspaces.}
    Realistic workspaces are substantially more complex: they contain more files, span a broader range of file types, and distribute
    task-relevant evidence throughout the directory hierarchy.
    In contrast, synthetic workspaces are highly curated, with most
    task-relevant information concentrated in a small set of identifiable files.
    }
    \label{fig:grounding-comparison}
\end{figure}

Recent work addresses this bottleneck through automated environment synthesis~\citep{song2026envscaler,tu2026scaleenv,cheng2026terminalworld,fan2026skillsynth,zhao2026nexforge,bai2026clawgym}.
However, these methods rely on synthetic environments or task-specific mock resources~\citep{song2026envscaler,tu2026scaleenv,bai2026clawgym}, yielding toy workspaces with limited file diversity and cross-file dependencies. This reduces the exploration demands that characterize real professional work in Figure~\ref{fig:grounding-comparison}.
Likewise, evaluation criteria not grounded in observable workspace facts risk introducing hallucinated or otherwise unsupported requirements, leading to unreliable judges~\citep{zhang2026agentv,shi2026facet}.

To address this challenge, we introduce \textsc{WorkForge}, a scalable framework for constructing complex, verifiable work-agent environments from real-world digital resources.
\textsc{WorkForge} is motivated by a simple premise: professional tasks and their verification are inseparable from the workspaces in which they are performed.
Workspace determines what an agent can explore, while observable workspace facts determine what can be asked and reliably verified.
To operationalize this premise, \textsc{WorkForge} formalizes expert workflows as task blueprints, retrieves relevant resources, and profiles the resulting workspaces to recover file roles, dependencies, and factual anchors.
These anchors jointly condition task generation and verifier design, keeping evaluation traceable to the workspace: programmatic checks verify deterministic requirements, while semantic rubrics assess open-ended deliverables against the same facts.
Autonomous solvers then execute these tasks to yield long-horizon trajectories for post-training.

\textsc{WorkForge} instantiates 16.7K verifiable environments across 40 professional domains, with workspaces comprising real artifacts, rich file types and cross-file dependencies.
Fine-tuning Qwen3.5-35B-A3B-Base~\citep{qwen2026qwen35} on the resulting trajectories improves performance across five benchmarks, including gains of 28.1 points on GDPval and 16.3 points on APEX.
Controlled studies display significant benefits with increasing volume and trajectory length, and confirm the efficacy of the proposed method.
Our contributions are as follows:
\begin{itemize}[leftmargin=*]
 \item We introduce \textsc{WorkForge}, a scalable framework that constructs verifiable environments from real-world resources, grounding task specifications and verification rubrics in audited workspaces.
  \item We instantiate 16.7K agentic environments across 40 professional domains, featuring complex filesystem and long-horizon solver trajectories.
  \item Empirical experiments demonstrate the efficacy of our method. Notably, fine-tuning on \textsc{WorkForge} data yields substantial gains across five diverse agent benchmarks, improving GDPval by 28.1 points and APEX by 16.3 points.
\end{itemize}

\section{\textsc{WorkForge}}
\label{sec:method}
\begin{figure*}[t]
  \centering
  \includegraphics[width=\textwidth]{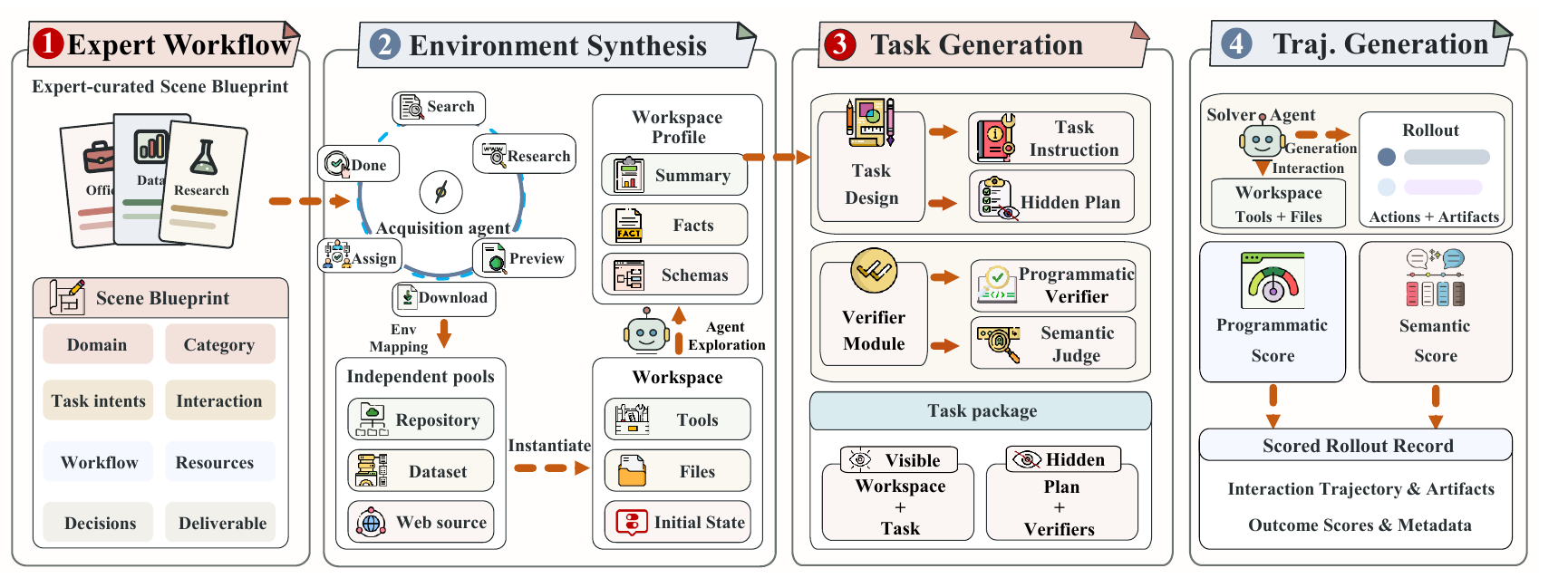}
  \caption{\textbf{Overview of \textsc{WorkForge} pipeline.} Conditioned on professional scene blueprints, \textsc{WorkForge} acquires and instantiates real-world resources into authentic workspaces, derives tasks and complementary verifiers, and collects scored trajectories through interaction.}
  \label{fig:WorkForge-method-pipeline}
\end{figure*}

To scale agent environments without compromising authenticity or verifiability,
we introduce \textsc{WorkForge}, a framework for translating professional
workflows into realistic, verifiable agent environments. \textsc{WorkForge}
first distills domain expertise into reusable workflow specifications. It then
instantiates each specification as a workspace populated with realistic
artifacts and analyzes the resulting workspace evidence to ground both task
construction and verification. Agents execute these tasks to produce
scored trajectories for training. 
Figure~\ref{fig:WorkForge-method-pipeline} summarizes this four-stage pipeline.

\subsection{Workflow Specification}
\label{sec:Workflow Specification}
To generate sophisticated professional tasks, we engage domain experts to distill their
workflows into reusable \emph{Scene Blueprints}, as these
workflows identify dependencies between initial conditions, domain distinctive operations,
and final deliverables. We define a collection of $M$ blueprints as
\begin{equation}
    \mathcal{B} = \{b_j\}_{j=1}^{M}, \qquad
    b_j = \left(
        \mathcal{W}_j,\mathcal{O}_j,\mathcal{R}_j,\mathcal{D}_j
    \right),
    \label{eq:scene-blueprint}
\end{equation}
where $\mathcal{W}_j$ specifies the workflow and its context; $\mathcal{O}_j$ defines the expected
deliverable; $\mathcal{R}_j$ specifies required sources and their dependencies;
and $\mathcal{D}_j$ records domain specific decisions.

For instance, a teacher can distill the workflow of testing
students on a unit and providing targeted explanations into
a scene blueprint. The blueprint specifies how to compare
student responses against a standard answer and use teaching
materials to group errors. It thus serves as a reusable seed
for constructing workspaces and generating tasks with
different assessment and teaching materials.

Each blueprint serves as a seed that connects expert intent to task generation.
Its resource requirements guide source acquisition. Simultaneously, its workflow,
decision points, and deliverable guide task generation using evidence from
the instantiated workspace.

\subsection{Environment Synthesis}
\label{sec:environment-synthesis}

Environment synthesis integrates two modules, environment mapping and environment profiling. The former one maps each scene blueprint to different
resource realizations. Complementing the mapping process, the latter one represents each environment's professional context,
resource characteristics, and observable facts.
For a scene blueprint $b_j$, the stage produces
\begin{equation}
    \operatorname{Env}(b_j)
        = \{(W_{jk},P_{jk})\}_{k=1}^{K_j},
    \label{eq:environment-synthesis}
\end{equation}
where $K_j$ is the number of retained environments, $W_{jk}$ is a workspace, and $P_{jk}$ is its profile.

\paragraph{Environment Mapping.}
Environment mapping expresses the one-to-many relationship between a scene
blueprint and its concrete environments. We acquire distinct
resource pools $\{R_{jk}\}_{k=1}^{K_j}$, each containing the files of a
repository and defining a resource mapping as
\begin{equation}
    W_{jk}=\phi_{R_{jk}}(b_j),
    \label{eq:workspace-instantiation}
\end{equation}
where $\phi_{R_{jk}}$ realizes $b_j$ using the selected resource pool.
An acquisition agent iteratively searches for sources, reads reference pages,
downloads resources, and previews their contents to assess relevance and
associate artifacts with the abstract source requirements $\mathcal R_j$.
Each workspace uses one non-empty pool and retains its file organization.

\paragraph{Environment Profiling.}
Environment profiling defines the description accompanying each
environment as an integral part of synthesis. We represent this profile as
\begin{equation}
    P_{jk}=\left(U_{jk},\mathcal F_{jk},A_{jk}\right),\qquad
    \mathcal F_{jk}=\left\{f_{jk}^{(\ell)}\right\}_{\ell=1}^{L_{jk}},
    \label{eq:workspace-context}
\end{equation}
where $U_{jk}$ summarizes the domain, file roles, and potential task
directions; $\mathcal F_{jk}$ contains workspace facts; and
$A_{jk}$ contains each artifact schemas and statistics.
In brief, the profile describes what makes each environment suitable for particular
professional tasks and supplies factual anchors for task construction and
subsequent solution verification.

\subsection{Task Generation}
\label{sec:task-generation}

Task generation produces executable task packages with aligned
task requirements and verification criteria. Task Design combines
the scene blueprint and environment profile to specify the
professional objective, required deliverables, and a hidden
solution plan. Verifier Design translates these requirements
into rubrics and complementary programmatic
and semantic verifiers.

\paragraph{Task Design.}
Given a scene blueprint $b_j$ and the profile $P_{jk}$,
the task designer jointly generates the instruction $q_{jk}$ which agent can view and hidden solution plan $h_{jk}$:
\begin{equation}
    (q_{jk},h_{jk})
    = \left(
        f_{\mathrm{task}}(b_j,P_{jk}),
        f_{\mathrm{plan}}(b_j,P_{jk})
      \right),
    \label{eq:workspace-conditioned-task}
\end{equation}
Together, the instruction and plan form an executable task grounded in the concrete
contents of the real-world workspace. Outputs with invalid paths or any unsupported components are regenerated.

\paragraph{Verification Module.}
We verify key factual claims in $h_{jk}$ against
the workspace $W_{jk}$ and record the resulting evidence as
$\widehat F_{jk}$. Along with the profile facts and task constraints,
this evidence defines an atomic rubric set
$\mathcal V_{jk}=\mathcal V_{jk}^{\mathrm{prog}}
\cup\mathcal V_{jk}^{\mathrm{sem}}$. After execution, programmatic criteria check the final artifacts
$Y_{jk}$:
\begin{equation}
    S_{jk}^{\mathrm{prog}}
    =
    \frac{1}{|\mathcal V_{jk}^{\mathrm{prog}}|}
    \sum_{v\in\mathcal V_{jk}^{\mathrm{prog}}}
    \mathbb I\!\left[v(W_{jk},Y_{jk})=1\right].
    \label{eq:programmatic-score}
\end{equation}
Semantic criteria and the task instruction $q_{jk}$ configure the LLM
judge $G_{jk}^{\mathrm{sem}}$, which assesses open-ended requirements
using criterion-specific scoring guides and normalizes each score
$s_v\in[0,10]$:
\begin{equation}
    S_{jk}^{\mathrm{sem}}
    =
    \frac{1}{|\mathcal V_{jk}^{\mathrm{sem}}|}
    \sum_{v\in\mathcal V_{jk}^{\mathrm{sem}}}
    \frac{s_v}{10}.
    \label{eq:semantic-score}
\end{equation}
Both verifiers are applied after execution, and their scores are
retained separately. The resulting task package separates
visible inputs from hidden solution plan and verifiers:
\begin{equation}
    \mathcal T_{jk}
    =
    \left(
      \underbrace{
        \left(W_{jk},q_{jk}\right)
      }_{\text{\footnotesize visible}},
      \underbrace{
        \left(
          P_{jk},h_{jk},\widehat F_{jk},\mathcal V_{jk},
          G_{jk}^{\mathrm{prog}},G_{jk}^{\mathrm{sem}}
        \right)
      }_{\text{\footnotesize hidden}}
    \right).
    \label{eq:task-package}
\end{equation}

\begin{figure}[t]
    \centering
    \begin{minipage}[t]{0.27\textwidth}
        \vspace{-10pt}
        \captionof{table}{Key statistics of \textsc{WorkForge} from task, workspace and trajectory three dimensions.}
        \label{tab:RACE-key-statistics}
        \centering
        \scriptsize
        \setlength{\tabcolsep}{2pt}
        \renewcommand{\arraystretch}{1.2}
        \begin{tabularx}{\linewidth}{@{}>{\raggedright\arraybackslash}Xr@{}}
            \toprule
            \textbf{Statistic} & \textbf{Value} \\
            \midrule
            \rowcolor{lightBlue}
            \multicolumn{2}{@{}l@{}}{\textbf{\textit{Task}}} \\
            \quad \# Task number & 16,665 \\
            \quad Avg. Rubrics & 21.9 \\
            \midrule
            \rowcolor{lightBlue}
            \multicolumn{2}{@{}l@{}}{\textbf{\textit{Workspace}}} \\
            \quad Avg. Files & 42.3 \\
            \quad Avg. Size & 279.0 KB \\
            \quad \# File Types & 60 \\
            \midrule
            \rowcolor{lightBlue}
            \multicolumn{2}{@{}l@{}}{\textbf{\textit{Trajectory}}} \\
            \quad Avg. Steps & 31.8 \\
            \quad Avg. Tool Calls & 33.8 \\
            \quad Avg. Tokens & 76,485 \\
            \bottomrule
        \end{tabularx}
    \end{minipage}
    \hfill
    \begin{minipage}[t]{0.71\textwidth}
        \vspace{-6pt}
        \centering
        \includegraphics[width=\linewidth]{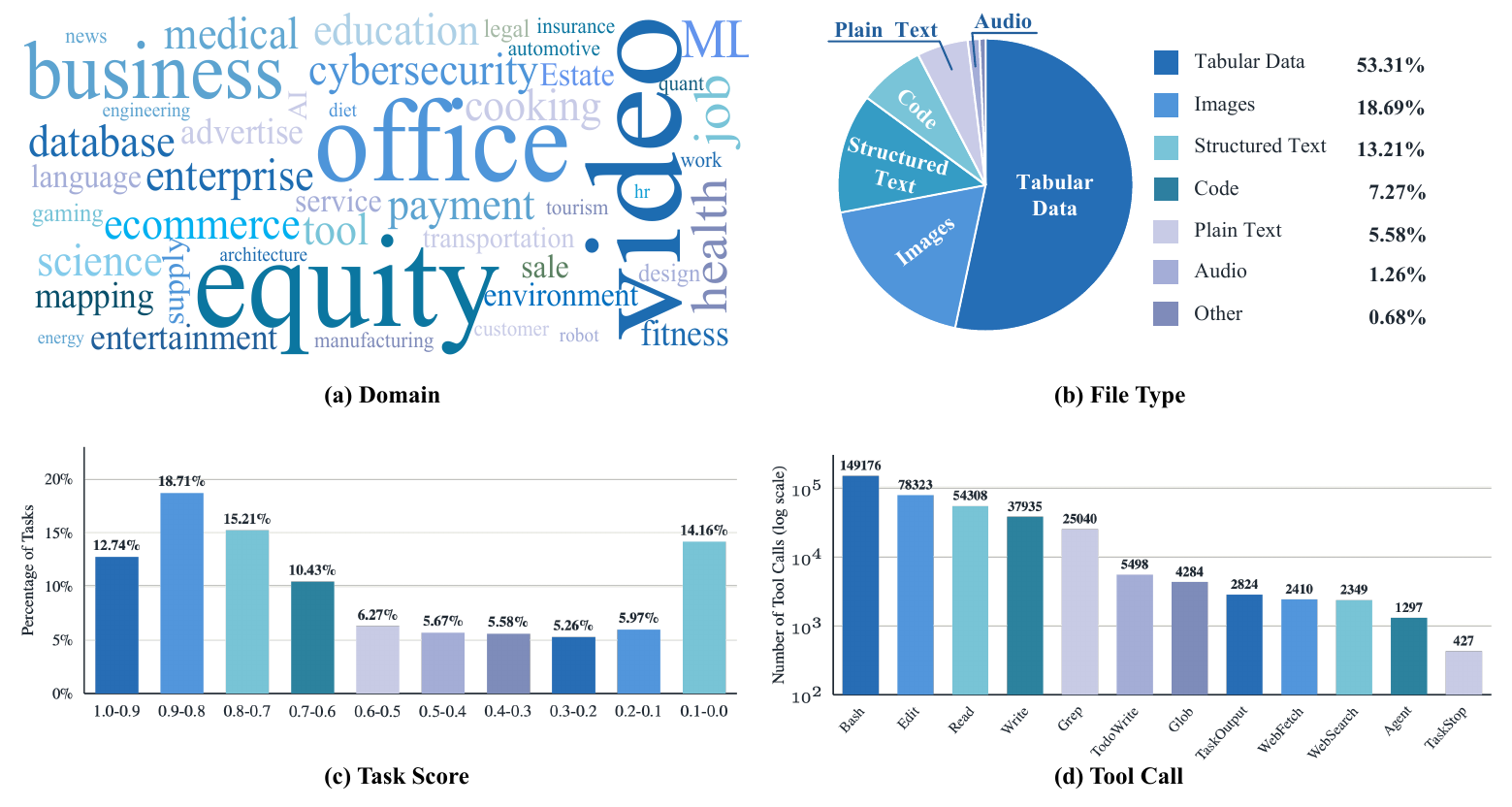}
        \vspace{-12pt}
        \caption{
        \textbf{Distributions of \textsc{WorkForge}.}
        Statistics across (a) Professional domains, (b) File types,
        (c) Task scores, and (d) Tool calls.
        }
        \label{fig:trajectory-distributions}
    \end{minipage}
\end{figure}
\subsection{Trajectory Generation}
\label{sec:trajectory-generation}
Trajectory generation
turns static task package into observed agent behavior.  For each
$\mathcal T_{jk}$, a solver agent $\pi_\theta$ receives only the visible pair
$(W_{jk},q_{jk})$ and interacts with the workspace until it submits its
deliverables or reaches the execution limit.  We denote this process as
\begin{equation}
    (\tau_{jk},Y_{jk})
    =\operatorname{Rollout}(\pi_\theta;q_{jk},W_{jk}),
    \label{eq:trajectory-rollout}
\end{equation}
where $\tau_{jk}$ records the ordered messages and tool interactions, and
$Y_{jk}$ denotes the output artifacts. 

After execution, the graders $G_{jk}^{\mathrm{prog}}$ and
$G_{jk}^{\mathrm{sem}}$ evaluate the final workspace and $Y_{jk}$, producing
the programmatic and semantic scores defined in
Eq.~\ref{eq:programmatic-score} and~\ref{eq:semantic-score}.  We retain the
complete execution as a scored trajectory record
\begin{equation}
    z_{jk}
    =\left(\tau_{jk},Y_{jk},S_{jk}^{\mathrm{prog}},
      S_{jk}^{\mathrm{sem}},\mu_{jk}\right),\qquad
    \mathcal D_{\mathrm{rollout}}=\left\{z_{jk}\right\}_{j,k},
    \label{eq:rollout-corpus}
\end{equation}
where $\mu_{jk}$ contains the model configuration.
Collectively, the four stages transform professional workflows encoded in scene blueprints
into independently grounded task packages and interaction
trajectories paired with verifiable execution outcomes. Table~\ref{tab:RACE-key-statistics}
reports the corpus-level scalar statistics, while
Figure~\ref{fig:trajectory-distributions} summarizes its domain coverage, file
types, task score, and tool call distribution.
Algorithm~\ref{alg:workforge-pipeline} in
Appendix~\ref{app:pipeline-prompts} summarizes the integral procedures from scene blueprints to trajectories.

\section{Experiments}
\label{sec:experiments}
\subsection{Experimental Setup}

\paragraph{Baselines.}
We evaluate models trained on \textsc{WorkForge}-16K against three groups of baselines.
\textbf{Frontier Proprietary Models:} Claude Opus 5.0~\citep{anthropic2026claudeopus5}, Claude Opus 4.8, GPT-5.6 Sol~\citep{openai2026gpt56}, and Gemini 3.7 Flash~\citep{doshi2026gemini37flash}.
\textbf{Open-source Models ($>$100B):} Qwen3-Coder~\citep{qwen2025qwen3coder}, Qwen3.5-397B-A17B~\citep{qwen2026qwen35}, DeepSeek-V3.2, DeepSeek-V4-Flash, DeepSeek-V4-Pro~\citep{deepseek2026v4}, Kimi-K2.5~\citep{kimi2026k25}, GLM-5.2~\citep{zai2026glm52}, MiniMax-M3~\citep{minimax2026m3}, and MiniMax-M2.7~\citep{minimax2026m27}.
\textbf{Open-source Models ($<$100B):} OpenThinkerAgent-32B~\citep{raoof2026openthoughts}, ClawGym-30A3~\citep{bai2026clawgym}, EnvScaler-Qwen3-8B~\citep{song2026envscaler}, Nemotron-Terminal-32B~\citep{pi2026data}, and Arctic-AWM-14B~\citep{wang2026agentworldmodel}.

\begin{table}[t]
    \caption{
    Performance (\%) of agent capability on WildClaw, ClawEval, GDPval, APEX, and Agent's Last Exam. The best results are in \textbf{bold} and the second-best results are \underline{underlined} within each group.
    Additional evaluation details are provided in Appendix~\ref{app:benchmark}.
    }

    \label{tab:main-results}
    \centering
    \scriptsize
    \setlength{\tabcolsep}{0.5pt}
    \setlength{\arrayrulewidth}{0.3pt}
    \renewcommand{\arraystretch}{1.1}
    \settowidth{\benchmarkcolwidth}{\textbf{WildClaw}}
    \settowidth{\benchmarkcandidatewidth}{\textbf{ClawEval}}
    \ifdim\benchmarkcandidatewidth>\benchmarkcolwidth
        \setlength{\benchmarkcolwidth}{\benchmarkcandidatewidth}
    \fi
    \settowidth{\benchmarkcandidatewidth}{\textbf{GDPval}}
    \ifdim\benchmarkcandidatewidth>\benchmarkcolwidth
        \setlength{\benchmarkcolwidth}{\benchmarkcandidatewidth}
    \fi
    \settowidth{\benchmarkcandidatewidth}{\textbf{APEX}}
    \ifdim\benchmarkcandidatewidth>\benchmarkcolwidth
        \setlength{\benchmarkcolwidth}{\benchmarkcandidatewidth}
    \fi
    \settowidth{\benchmarkcandidatewidth}{\textbf{ALE}}
    \ifdim\benchmarkcandidatewidth>\benchmarkcolwidth
        \setlength{\benchmarkcolwidth}{\benchmarkcandidatewidth}
    \fi
    \begin{tabular}{@{}l!{\vrule width \arrayrulewidth}*{9}{w{c}{\benchmarkcolwidth}}@{}}
        \toprule
        \multirow{2}{*}{\textbf{Model}} &
        \multicolumn{2}{!{\vrule width 0.3pt}c!{\vrule width 0.3pt}}{\textbf{WildClaw}} &
        \multicolumn{2}{c!{\vrule width 0.3pt}}{\textbf{ClawEval}} &
        \multicolumn{1}{c!{\vrule width 0.3pt}}{\textbf{GDPval}} &
        \multicolumn{2}{c!{\vrule width 0.3pt}}{\textbf{APEX}} &
        \multicolumn{2}{c}{\textbf{ALE}} \\
        
        \cmidrule(lr){2-3}
        \cmidrule(lr){4-5}
        \cmidrule(lr){6-6}
        \cmidrule(lr){7-8}
        \cmidrule(lr){9-10}
        
        &
        \shortstack{\textbf{Score}} &
        \multicolumn{1}{c!{\vrule width 0.3pt}}{\textbf{Pass@1}} &
        \textbf{Score} &
        \multicolumn{1}{c!{\vrule width 0.3pt}}{\textbf{Pass\symbol{94}3}  } &
        \multicolumn{1}{c!{\vrule width 0.3pt}}{\textbf{Score}} &
        \shortstack{\textbf{Score}} &
        \multicolumn{1}{c!{\vrule width 0.3pt}}{\textbf{Pass@1}} &
        \textbf{Score} &
        \textbf{Pass@1} \\
        \specialrule{0.8pt}{0.5pt}{2pt}
        \noalign{\global\arrayrulewidth=0pt}
        \multicolumn{10}{@{}c@{}}{\textbf{\textit{Frontier Proprietary Models}}} \\
        \noalign{\global\arrayrulewidth=0.3pt}
        \midrule
        Claude Opus 5.0 & \textbf{64.5} & \textbf{32.4} & \textbf{82.3} & \textbf{74.2} & \textbf{91.0} & \textbf{58.5} & \textbf{42.7} & \textbf{48.3} & \textbf{26.7} \\
        Claude Opus 4.8 & \underline{64.0} & \underline{30.5} & \underline{81.5} & \underline{72.1} & 89.1 & 46.2 & 34.0 & 46.2 & 24.8 \\
        GPT-5.6 Sol & 53.5 & 29.5 & 77.3 & 68.9 & \underline{90.1} & \underline{51.8} & \underline{37.5} & \textbf{48.3} & \textbf{26.7} \\
        Gemini 3.7 Flash & 52.6 & 18.1 & 69.2 & 60.0 & 86.1 & 51.1 & 36.9 & \underline{46.9} & \underline{25.7} \\
        \specialrule{0.8pt}{1.0pt}{2pt}
        \noalign{\global\arrayrulewidth=0pt}
        \multicolumn{10}{@{}c@{}}{\textbf{\textit{Open-source Models ($>$100B)}}} \\
        \noalign{\global\arrayrulewidth=0.3pt}
        \midrule
        Qwen3-Coder & 32.7 & 15.2 & 37.3 & 23.4 & 67.7 & 11.4 & 2.7 & 21.1 & 5.7 \\
        Qwen3.5-397B-A17B & 38.8 & 17.1 & 60.8 & 53.0 & 79.7 & 27.4 & 12.9 & 31.1 & 11.4 \\
        DeepSeek-V3.2 & 28.2 & 15.2 & 58.3 & 49.2 & 74.2 & 17.6 & 6.3 & 26.3 & 9.5 \\
        DeepSeek-V4-Flash & \underline{59.2} & 27.6 & \textbf{74.4} & \textbf{66.4} & 83.6 & \underline{46.3} & 30.0 & 42.3 & 22.9 \\
        DeepSeek-V4-Pro & \textbf{60.7} & \textbf{31.4} & 70.4 & 61.6 & \textbf{88.4} & \textbf{48.5} & \textbf{32.1} & \underline{45.3} & \underline{23.2} \\
        Kimi-K2.5 & 41.1 & 14.3 & 56.7 & 44.4 & 78.5 & 24.1 & 11.0 & 32.5 & 11.4 \\
        GLM-5.2 & 58.2 & 28.6 & \underline{73.1} & \underline{63.8} & 85.2 & 45.7 & \underline{30.4} & \textbf{47.9} & \textbf{24.8} \\
        MiniMax-M3 & 56.8 & 28.6 & 70.2 & 55.6 & \underline{86.8} & 41.2 & 26.0 & 34.6 & 17.1 \\
        MiniMax-M2.7 & 41.2 & 16.2 & 60.8 & 48.7 & 81.9 & 25.3 & 11.5 & 28.4 & 9.5 \\
        \specialrule{0.8pt}{1.0pt}{2pt}
        \noalign{\global\arrayrulewidth=0pt}
        \multicolumn{10}{@{}c@{}}{\textbf{\textit{Open-source Models ($<$100B)}}} \\
        \noalign{\global\arrayrulewidth=0.3pt}
        \midrule
        OpenThinker-32B & \textbf{16.7} & 5.7 & \underline{31.2} & \underline{17.2} & 21.2 & \textbf{6.8} & \textbf{3.1} & \underline{2.0} & 0.0 \\
        ClawGym-30A3 & \underline{14.9} & \underline{6.7} & \textbf{35.4} & \textbf{19.4} & \textbf{35.0} & 2.5 & 0.4 & \textbf{9.1} & \textbf{2.9} \\
        EnvScaler-8B & 13.7 & 3.8 & 18.7 & 8.2 & 15.0 & \underline{4.9} & \underline{0.8} & 0.7 & 0.0 \\
        Nemotron-Terminal-32B & 8.8 & 2.9 & 12.9 & 1.8 & 5.8 & 1.8 & 0.6 & 1.2 & 0.0 \\
        Agent World Model 14B & 13.1 & \textbf{7.6} & 26.0 & 16.4 & \underline{24.5} & 3.6 & \underline{0.8} & 1.1 & \underline{0.8} \\
        \specialrule{0.8pt}{1.0pt}{2pt}
        \noalign{\global\arrayrulewidth=0pt}
        \multicolumn{10}{@{}c@{}}{\textbf{\textit{Ours}}} \\
        \noalign{\global\arrayrulewidth=0.3pt}
        \midrule
        Qwen3.5-35B-A3B-Base & 18.6 & 7.6 & 35.4 & 19.9 & 45.5 & 5.0 & 0.6 & 11.4 & 3.8 \\
        \hspace{1em}$+$ WorkForge-A3B & 29.9 & \underline{20.0} & 53.3 & 41.8 & 73.6 & 21.3 & 9.2 & 28.7 & 12.4 \\
        \addlinespace[0.5pt]
        Qwen3.5-27B & \underline{30.5} & 18.1 & \underline{61.9} & \underline{52.0} & \underline{79.4} & \underline{25.7} & \underline{12.7} & \underline{32.7} & \underline{14.3} \\
         \hspace{1em}$+$ WorkForge-27B & \textbf{38.5} & \textbf{23.8} & \textbf{65.4} & \textbf{55.7} & \textbf{82.4} & \textbf{29.9} & \textbf{16.3} & \textbf{33.9} & \textbf{17.1} \\
        \bottomrule
    \end{tabular}
\end{table}

\paragraph{Benchmarks.}
We evaluate work-agent capability on five popular benchmarks.
\textbf{WildClawBench}~\citep{ding2026wildclawbench} and \textbf{Claw-Eval}~\citep{ye2026claweval} emphasize long-horizon execution, multi-file interaction, and reliable tool use under repeated trials.
\textbf{GDPval}~\citep{patwardhan2026gdpval} and \textbf{APEX-Agents (APEX)}~\citep{vidgen2026apex} focus on professional deliverables and workplace tool orchestration, while \textbf{Agents' Last Exam (ALE)}~\citep{sun2026agentslastexam} tests extended autonomous problem solving with verifiable outcomes.
Additional evaluation details are provided in Appendix~\ref{app:benchmark}.

\paragraph{Implementation Details.}
We apply supervised fine-tuning on Qwen3.5-35B-A3B-Base and Qwen3.5-27B.
Both models are trained for five epochs with a global batch size of $64$ and a maximum
sequence length of $256$K tokens. 
We adopt a learning rate of $1\times10^{-5}$ with cosine decay and 5\% warmup.
More implementation illustrations are in Appendix~\ref{app:experimental-details}.

\subsection{Main Results}

\paragraph{Superior performance of \textsc{WorkForge} over other baselines.}
Table~\ref{tab:main-results} reports the capacity of general agents across state-of-the-art competitors.
We make several key observations: (1) \textsc{WorkForge} delivers consistently competitive results across all benchmarks, establishing strong performance. 
Notably, \textsc{WorkForge}-A3B achieves the highest GDPval Score among prior specialized sub‑100B baselines, i.e., 73.6, surpassing the previous SOTA, Clawgym-30A3, by a substantial margin of 38.6 points. Similar trends are observed on WildClaw, ClawEval, APEX, and ALE. 

(2) Importantly, these leads do not shrink under strict pass rate: 
\textsc{WorkForge}-A3B reaches 12.4\% Pass@1 on ALE, compared with 2.9\% of ClawGym-30A3 which rarely complete these challenge tasks. 
(3) \textsc{WorkForge}-A3B also delivers a substantial improvement over pretrained model Qwen3.5-35B-A3B-Base. 
These gains reflect \textsc{WorkForge}'s design principle: solving each task requires the agent to explore complex, real-world file structures and integrate evidence across multiple files into its deliverables. This capability generalizes well on challenging benchmarks.

\paragraph{\textsc{WorkForge}-27B advances capability toward the frontier.}
\textsc{WorkForge}-27B demonstrates that the supervision produced by our pipeline complements existing agent post-training, improving Qwen3.5-27B across all five benchmarks, with WildClaw, APEX, and GDPval Scores rising by 8.0, 4.2, and 3.0 points, respectively.
Notably, on GDPval, \textsc{WorkForge}-27B reaches 82.4, surpassing Qwen3.5-397B-A17B (79.7) despite nearly 15$\times$ fewer total parameters, and narrowing the gap to frontier systems such as Claude Opus 4.8 (89.1) and GPT-5.6 Sol (90.1).

\begin{figure}[t]
    \centering
    \includegraphics[width=\linewidth]{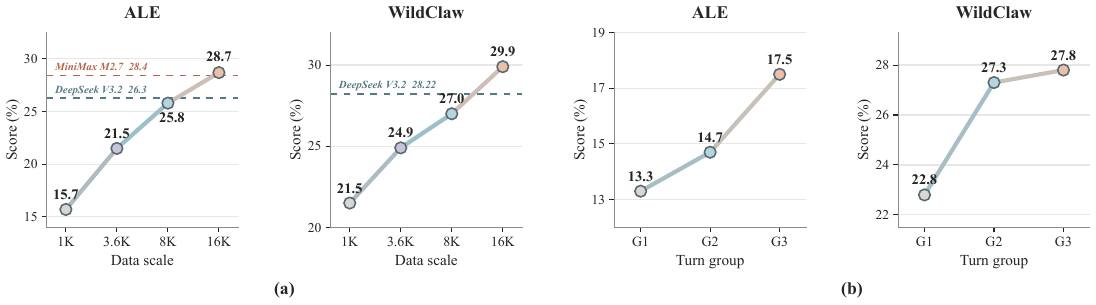}
    \caption{Scaling analysis of \textsc{WorkForge}. Performance on ALE and WildClaw with increasing (a) Training data volume and (b) Interaction length.}
    \label{fig:scaling-analysis}
\end{figure}

\subsection{Ablation Studies}

\begin{wrapfigure}{r}{0.46\linewidth}
    \vspace{-1.8\baselineskip}
    \centering
    \includegraphics[width=\linewidth]{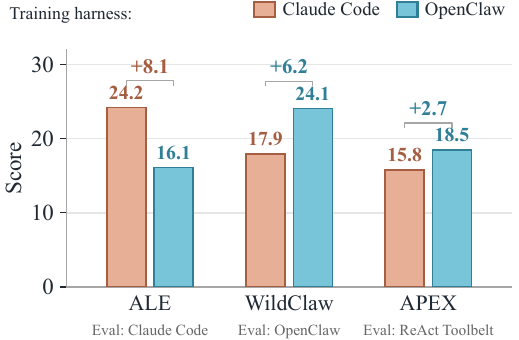}
    \caption{\textbf{Effect of trajectory harness.} We compare two training groups, each using 8k trajectories generated by Claude Code or OpenClaw, respectively.}
    \label{fig:harness-ablation}
    \vspace{-0.6\baselineskip}
\end{wrapfigure}
\paragraph{Trajectory Harness.}
We further compare models trained on trajectories generated by two agent harnesses, Claude Code~\citep{anthropic2025claudecode} and OpenClaw~\citep{steinberger2026openclaw}. In Figure~\ref {fig:harness-ablation}, the model trained on Claude Code trajectories achieves a higher ALE score when evaluated with Claude Code (24.2 vs.\ 16.1). On WildClaw, this ordering reverses: the model trained on OpenClaw trajectories leads by 6.2 points under the OpenClaw evaluation harness. The OpenClaw-trained model also performs better on APEX, although its advantage narrows to 2.7 points under the official ReAct Toolbelt.
These results indicate a degree of harness-specific transfer: aligning the training and evaluation harnesses is associated with improved performance, potentially because the model acquires harness-specific tool-use conventions in addition to task knowledge.

\begin{wraptable}{r}{0.46\linewidth}
    \vspace{-0.8\baselineskip}
    \centering
    \caption{Ablation of verification-based trajectory selection. Each variant uses trajectories selected by the semantic score, the programmatic score, or their combination.}
    \label{tab:verification-ablation}
    \footnotesize
    \setlength{\tabcolsep}{3.5pt}
    \renewcommand{\arraystretch}{1.0}
    \begin{tabularx}{\linewidth}{@{}Xrrr@{}}
        \toprule
        \multirow{1}{*}{} & \multicolumn{2}{c}{ALE} & APEX \\
        \cmidrule(lr){2-3}
        \cmidrule(lr){4-4}
        & Score & Pass@1 & Score \\
        \midrule
        \rule[-3pt]{0pt}{12pt}LLM only
        & 15.62
        & 4.8
        & 11.18 \\
        \rule[-3pt]{0pt}{12pt}Programmatic
        & 14.89
        & 4.8
        & 10.92 \\
        \rule[-3pt]{0pt}{12pt}Hybrid
        & \textbf{22.11}
        & \textbf{11.4}
        & \textbf{15.39} \\
        \bottomrule
    \end{tabularx}
\end{wraptable}
\paragraph{Verification Mechanism.}
Recall that \textsc{WorkForge} evaluates task deliverables with two components: programmatic criteria check exact values and observable workspace states, and LLM-based criteria assess content quality.
To isolate their effects on trajectory selection, we construct three 1K-trajectory training sets from the same pool of candidate tasks.
The programmatic-only and LLM-only variants select trajectories whose corresponding score is at least 0.7, while the hybrid variant applies the same threshold to the equally weighted mean of the two scores.
All variants use the same training-set size and fine-tuning configuration.
As shown in Table~\ref{tab:verification-ablation}, the hybrid strategy outperforms either variant.
This indicates that combining the two verification signals yields effective supervision.

\paragraph{Workspace Authenticity.}
We instantiate matched workspaces from the same scene blueprints and vary only whether their resources are retrieved from the web or synthesized by an LLM.
Figure~\ref{fig:grounding-comparison} reveals a sharp structural gap: workspaces built from real-world resources contain 5.4$\times$ more files and 2.4$\times$ more file types, with their average footprint increasing from 21.5\,KB to 279\,KB.
Real-world acquisition therefore enlarges both the scale and heterogeneity of the workspace, providing a richer artifact and evidence space for task construction and agent exploration.

\begin{figure*}[t!]
    \centering
    \includegraphics[width=\linewidth]{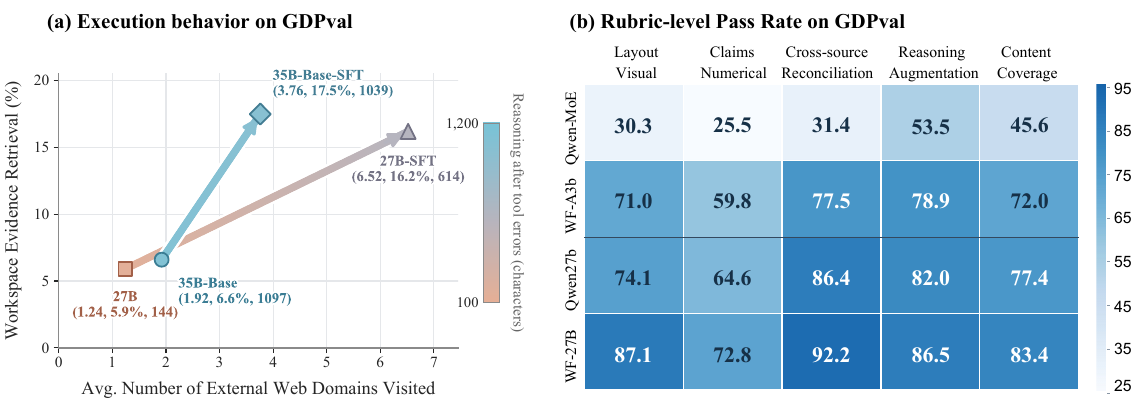}
    \caption{Execution behavior and rubric-level performance on GDPval.
        (a) The x-axis presents the mean number of distinct external web domains visited per task; the y-axis presents evidence retrieval within the workspace as a percentage of tool calls. 
        Arrows connect each base model to its fine-tuned counterpart; color records reasoning length after tool errors.
        (b) Pass rate across five rubric categories. Details about rubric categories are described in Table~\ref {tab:gdpval-rubric-categories}.}
    \label{fig:behavior-analysis}
\end{figure*}

\subsection{Scaling Analysis}

\paragraph{Scaling with Data Volume.} 
To investigate how training data volume affects work-agent performance, we train Qwen3.5-35B-A3B-Base on 1K, 3.6K, 8K, and 16K trajectories under the same recipe.
As shown in Figure~\ref{fig:scaling-analysis}~(
a), performance improves monotonically with data volume.
The model trained on 16K trajectories achieves scores of 28.7 on ALE and 29.9 on WildClaw, outperforming its 1K-trajectory counterpart by 13.0 and 8.4 points, respectively.
At this scale, the model also surpasses DeepSeek V3.2 (26.3) and MiniMax M2.7 (28.4) on ALE, as well as DeepSeek V3.2 (28.2) on WildClaw.
Increasing data volume exposes the model to a broader range of workspace structures, cross-file dependencies, and execution paths.
Broader coverage promotes transferable capabilities for workspace exploration and deliverable production.

\paragraph{Scaling with Interaction Horizon.}
Beyond data volume, we explore how interaction horizon shapes agent behavior.
We construct three subsets of 1K Claude Code trajectories, grouped by trajectory length: 0--20 turns (G1), 21--40 turns (G2), and 41--60 turns (G3).
As shown in Figure~\ref{fig:scaling-analysis}~(b), performance increases with trajectory length on both benchmarks.
The model trained on G3 achieves scores of 17.5 on ALE and 27.8 on WildClaw, exceeding its G1 counterpart by 4.2 and 5.0 points, respectively.
This improvement is accompanied by more sustained interaction with the environment.
On WildClaw, the G3-trained model averages 24.0 productive tool calls, compared with 11.7 for the G1-trained model, and performs 38.3 distinct actions, compared with 17.0.
Table~\ref{tab:turn-scaling-behavior} further shows that the G3-trained model uses a broader range of tool types and invokes workspace skills more frequently.
These results suggest that longer training trajectories expose the model to more complete sequences of dependency tracing, evidence gathering, and deliverable construction, thereby promoting the sustained interaction required in complex workspaces.

\subsection{Behavior Analysis}
We analyze the trajectory behavior of \textsc{WorkForge}-27B.
Figure~\ref{fig:behavior-analysis}~(a) reveals a shift toward exploration behavior: the fine-tuned model visits 6.52 external web domains per task (5.3$\times$ the base model's average) and lifts targeted evidence retrieval from 5.9\% to 16.2\% of tool calls.
After a failed tool call, reasoning length grows from 144 to 614 characters (over fourfold), reflecting deliberate diagnosis and replanning rather than immediate retries.
These shifts carry through to final deliverables: Figure~\ref{fig:behavior-analysis}~(b) shows category-score gains of 4.5--13 points across five GDPval-derived categories.
Consistent with this shift toward exploration behavior, the largest gains appear in formatting and visual presentation (+13 pp) and numerical and calculation accuracy (+8.2 pp), which rely on thorough investigation and careful verification.

\section{Related Work}
\label{sec:related-work}

\paragraph{Agent Environment Synthesis.}
Automated environment construction expands the supply of executable settings for agent training.
EnvScaler and Agent World Model synthesize tool interfaces and stateful backends \citep{song2026envscaler,wang2026agentworldmodel}, while Agent-World incorporates real-service metadata and external resources \citep{dong2026agentworld}.
Skill and intent-driven pipelines construct tasks and supporting workspaces for terminal and general-assistant workflows \citep{cheng2026terminalworld,bai2026clawgym}.
NexForge compiles capability requirements into task directives, then retrieves or constructs the resources needed for execution \citep{zhao2026nexforge}.
\textsc{WorkForge} instead instantiates and profiles realistic workspaces before specifying concrete tasks.
Preserving repository and dataset organization grounds professional tasks in heterogeneous materials, existing file relationships, and observable evidence.

\paragraph{Long-Horizon Agent Data Synthesis.}
Agentic data synthesis expands multi-step execution through tool interactions \citep{prabhakar2025apigenmt,xu2025toucan}, task composition \citep{shi2025taskcraft,xie2025agentsynth}, and exploration-based task discovery \citep{ramrakhya2025autoplay,chen2026dive,wang2026trajectory2task}.
Nemotron-Terminal studies generation, filtering, and long-context training \citep{pi2026data}, while OpenThoughts-Agent examines task sources and diversity to train OpenThinkerAgent-32B \citep{raoof2026openthoughts}.
Executable-state alignment \citep{shi2026facet} and rubric-based artifact assessment \citep{starace2025paperbench,zhuge2025agentjudge} connect trajectory quality to task outcomes.
\textsc{WorkForge} generates long-horizon supervision through sustained exploration, cross-file reasoning, and artifact production in content-rich workspaces.
Task requirements and verification criteria share the same workspace evidence, enabling complementary programmatic and semantic evaluation to select trajectories for agent post-training.

\section{Conclusion, Limitations and Future Work}
\paragraph{Conclusion}
We introduce \textsc{WorkForge}, a scalable framework for constructing complex, verifiable work-agent environments from real-world resources. Its core principle is to use observable workspace facts as shared anchors for task construction and outcome verification. 
Starting from expert workflows, \textsc{WorkForge} retrieves and profiles real-world resources, then derives task specifications, solution plans, and complementary programmatic and semantic verifiers from the resulting factual anchors. 
The pipeline instantiated 16.7K environments across 40 professional domains, which improves the capability of Qwen3.5-35B-A3B-Base, including gains of 28.1 points on GDPval and 16.3 points on APEX. 
Our analysis indicates significant gains with increasing data volume and trajectory length, while confirming the efficacy of verification mechanism.

\paragraph{Limitations and Future Work}
\textsc{WorkForge} depends on expert workflows and accessible realistic resources, limiting comparable grounding in proprietary or sensitive domains. Because workspace profiles, tasks, and verifiers are constructed automatically, their reliability depends on the completeness and accuracy of workspace profiling. Moreover, the current verifiers emphasize final deliverables, receive limited expert calibration, and do not directly assess intermediate agent behavior. Future work should broaden workflow and resource coverage while developing verification that is aware of processing.

\subsection*{AI Use Statement}
Generative AI tools were used to assist and polish manuscript writing. They were also used within \textsc{WorkForge} to generate synthetic task packages and training trajectories, as detailed in the method and experimental sections. The authors review all AI-assisted outputs and take responsibility for the final content of this work.

\subsection*{Ethics Statement}
This paper introduces a scalable framework for constructing complex, verifiable work-agent environments from real-world digital resources.
The methods and findings presented are intended solely for research purposes; our study does not pertain to malicious applications, unintended uses, or issues related to fairness, privacy, security, crowdsourcing, or human subject research.

\bibliography{references}

@inproceedings{zhou2024webarena,
  author = {Zhou, Shuyan and Xu, Frank F. and Zhu, Hao and Zhou, Xuhui and Lo, Robert and Sridhar, Abishek and Cheng, Xianyi and Ou, Tianyue and Bisk, Yonatan and Fried, Daniel and Alon, Uri and Neubig, Graham},
  title = {{WebArena}: A Realistic Web Environment for Building Autonomous Agents},
  booktitle = {International Conference on Learning Representations},
  year = {2024},
  url = {https://proceedings.iclr.cc/paper_files/paper/2024/file/4410c0711e9154a7a2d26f9b3816d1ef-Paper-Conference.pdf}
}

@inproceedings{trivedi2024appworld,
  author = {Trivedi, Harsh and Khot, Tushar and Hartmann, Mareike and Manku, Ruskin and Dong, Vinty and Li, Edward and Gupta, Shashank and Sabharwal, Ashish and Balasubramanian, Niranjan},
  title = {{AppWorld}: A Controllable World of Apps and People for Benchmarking Interactive Coding Agents},
  booktitle = {Proceedings of the 62nd Annual Meeting of the Association for Computational Linguistics},
  pages = {16022--16076},
  year = {2024},
  doi = {10.18653/v1/2024.acl-long.850}
}

@inproceedings{wang2026trajectory2task,
  author = {Wang, Ziyi and Lu, Yuxuan and Zhang, Yimeng and Chen, Pei and Dong, Ziwei and Huang, Jing and Gesi, Jiri and Tang, Xianfeng and Luo, Chen and Liu, Qun and Sang, Yisi and Lu, Hanqing and Li, Manling and Lai, Jin and Wang, Dakuo},
  title = {{Trajectory2Task}: Training Robust Tool-Calling Agents with Synthesized Yet Verifiable Data for Complex User Intents},
  booktitle = {Proceedings of the 64th Annual Meeting of the Association for Computational Linguistics},
  pages = {44021--44044},
  year = {2026},
  doi = {10.18653/v1/2026.acl-long.2037}
}

@misc{tang2026workspacebench,
  author = {Tang, Zirui and Zhou, Xuanhe and Liu, Yumou and Li, Linchun and Wu, Yukai and Wang, Weizheng and Huang, Hongzhang and Zhou, Wei and Zhou, Jun and Song, Jiachen and Yu, Shaoli and Wang, Jinqi and Zhou, Zihang and Zhou, Hongyi and Lv, Yuting and Li, Jinyang and Liu, Jiashuo and Chen, Ruoyu and Liu, Chunwei and Li, GuoLiang and Kang, Jihua and Wu, Fan},
  title = {{Workspace-Bench} 1.0: Benchmarking {AI} Agents on Workspace Tasks with Large-Scale File Dependencies},
  year = {2026},
  eprint = {2605.03596},
  archivePrefix = {arXiv},
  primaryClass = {cs.AI},
  url = {https://arxiv.org/abs/2605.03596}
}

@article{starace2025paperbench,
  author = {Starace, Giulio and Jaffe, Oliver and Sherburn, Dane and Aung, James and Chan, Jun Shern and Maksin, Leon and Dias, Rachel and Mays, Evan and Kinsella, Benjamin and Thompson, Wyatt and Heidecke, Johannes and Glaese, Amelia and Patwardhan, Tejal},
  title = {{PaperBench}: Evaluating {AI}'s Ability to Replicate {AI} Research},
  journal = {arXiv preprint arXiv:2504.01848},
  year = {2025},
  doi = {10.48550/arXiv.2504.01848}
}

@inproceedings{drouin2024workarena,
  author = {Drouin, Alexandre and Gasse, Maxime and Caccia, Massimo and Laradji, Issam H. and Del Verme, Manuel and Marty, Tom and Vazquez, David and Chapados, Nicolas and Lacoste, Alexandre},
  title = {{WorkArena}: How Capable Are Web Agents at Solving Common Knowledge Work Tasks?},
  booktitle = {Proceedings of the 41st International Conference on Machine Learning},
  volume = {235},
  series = {Proceedings of Machine Learning Research},
  pages = {11642--11662},
  year = {2024},
  publisher = {PMLR},
  url = {https://proceedings.mlr.press/v235/drouin24a.html}
}

@inproceedings{boisvert2024workarenapp,
  author = {Boisvert, L\'{e}o and Thakkar, Megh and Gasse, Maxime and Caccia, Massimo and De Chezelles, Thibault Le Sellier and Cappart, Quentin and Chapados, Nicolas and Lacoste, Alexandre and Drouin, Alexandre},
  title = {{WorkArena++}: Towards Compositional Planning and Reasoning-Based Common Knowledge Work Tasks},
  booktitle = {Advances in Neural Information Processing Systems},
  volume = {37},
  pages = {5996--6051},
  year = {2024},
  doi = {10.52202/079017-0195}
}

@inproceedings{huang2024mlagentbench,
  author = {Huang, Qian and Vora, Jian and Liang, Percy and Leskovec, Jure},
  title = {{MLAgentBench}: Evaluating Language Agents on Machine Learning Experimentation},
  booktitle = {Proceedings of the 41st International Conference on Machine Learning},
  volume = {235},
  series = {Proceedings of Machine Learning Research},
  pages = {20271--20309},
  year = {2024},
  publisher = {PMLR},
  url = {https://proceedings.mlr.press/v235/huang24y.html}
}

@inproceedings{prabhakar2025apigenmt,
  author = {Prabhakar, Akshara and Liu, Zuxin and Zhu, Ming and Zhang, Jianguo and Awalgaonkar, Tulika Manoj and Wang, Shiyu and Liu, Zhiwei and Chen, Haolin and Hoang, Thai and Niebles, Juan Carlos and Heinecke, Shelby and Yao, Weiran and Wang, Huan and Savarese, Silvio and Xiong, Caiming},
  title = {{APIGen-MT}: Agentic Pipeline for Multi-Turn Data Generation via Simulated Agent-Human Interplay},
  booktitle = {Advances in Neural Information Processing Systems},
  volume = {38},
  year = {2025},
  url = {https://proceedings.neurips.cc/paper_files/paper/2025/file/5e3661f7fe4c8ac5652d62eb3d3c96ea-Paper-Datasets_and_Benchmarks_Track.pdf}
}

@misc{xu2025toucan,
  author = {Xu, Zhangchen and Soria, Adriana Meza and Tan, Shawn and Roy, Anurag and Agrawal, Ashish Sunil and Poovendran, Radha and Panda, Rameswar},
  title = {{TOUCAN}: Synthesizing 1.5M Tool-Agentic Data from Real-World {MCP} Environments},
  year = {2025},
  eprint = {2510.01179},
  archivePrefix = {arXiv},
  primaryClass = {cs.LG},
  url = {https://arxiv.org/abs/2510.01179}
}

@inproceedings{zhuge2025agentjudge,
  author = {Zhuge, Mingchen and Zhao, Changsheng and Ashley, Dylan R. and Wang, Wenyi and Khizbullin, Dmitrii and Xiong, Yunyang and Liu, Zechun and Chang, Ernie and Krishnamoorthi, Raghuraman and Tian, Yuandong and Shi, Yangyang and Chandra, Vikas and Schmidhuber, J\"{u}rgen},
  title = {Agent-as-a-Judge: Evaluate Agents with Agents},
  booktitle = {Proceedings of the 42nd International Conference on Machine Learning},
  volume = {267},
  series = {Proceedings of Machine Learning Research},
  pages = {80569--80611},
  year = {2025},
  publisher = {PMLR},
  url = {https://proceedings.mlr.press/v267/zhuge25a.html}
}

@misc{song2026envscaler,
  author = {Song, Xiaoshuai and Chang, Haofei and Dong, Guanting and Zhu, Yutao and Wen, Ji-Rong and Dou, Zhicheng},
  title = {{EnvScaler}: Scaling Tool-Interactive Environments for {LLM} Agent via Programmatic Synthesis},
  year = {2026},
  eprint = {2601.05808},
  archivePrefix = {arXiv},
  primaryClass = {cs.CL},
  doi = {10.48550/arXiv.2601.05808},
  url = {https://arxiv.org/abs/2601.05808}
}

@misc{wang2026agentworldmodel,
  author = {Wang, Zhaoyang and Xu, Canwen and Liu, Boyi and Wang, Yite and Han, Siwei and Yao, Zhewei and Yao, Huaxiu and He, Yuxiong},
  title = {Agent World Model: Infinity Synthetic Environments for Agentic Reinforcement Learning},
  year = {2026},
  eprint = {2602.10090},
  archivePrefix = {arXiv},
  primaryClass = {cs.AI},
  doi = {10.48550/arXiv.2602.10090},
  url = {https://arxiv.org/abs/2602.10090}
}

@misc{dong2026agentworld,
  author = {Dong, Guanting and Lu, Junting and Huang, Junjie and Zhong, Wanjun and Liu, Longxiang and Huang, Shijue and Li, Zhenyu and Zhao, Yang and Song, Xiaoshuai and Li, Xiaoxi and Jin, Jiajie and Zhu, Yutao and Wang, Hanbin and Lei, Fangyu and Luo, Qinyu and Chen, Mingyang and Chen, Zehui and Feng, Jiazhan and Wen, Ji-Rong and Dou, Zhicheng},
  title = {{Agent-World}: Scaling Real-World Environment Synthesis for Evolving General Agent Intelligence},
  year = {2026},
  eprint = {2604.18292},
  archivePrefix = {arXiv},
  primaryClass = {cs.AI},
  doi = {10.48550/arXiv.2604.18292},
  url = {https://arxiv.org/abs/2604.18292}
}

@misc{cheng2026terminalworld,
  author = {Cheng, Zihao and Wang, Hongru and Liu, Zeming and Wang, Xinyi and Zhu, Xiangrong and Guo, Yuhang and Lin, Wei and Pan, Jeff Z. and Wang, Yunhong},
  title = {{Terminal-World}: Scaling Terminal-Agent Environments via Agent Skills},
  year = {2026},
  eprint = {2605.20876},
  archivePrefix = {arXiv},
  primaryClass = {cs.CL},
  doi = {10.48550/arXiv.2605.20876},
  url = {https://arxiv.org/abs/2605.20876}
}

@misc{zhao2026nexforge,
  author = {Zhao, Jiarong and Lei, Zhikai and Xi, Zhiheng and Zheng, Rui and Yan, Hang and Zhou, Jie and Chen, Qin and He, Liang},
  title = {{NexForge}: Scaling Agent Capabilities through Requirement-Driven Task Synthesis for {LLMs}},
  year = {2026},
  eprint = {2607.14186},
  archivePrefix = {arXiv},
  primaryClass = {cs.SE},
  doi = {10.48550/arXiv.2607.14186},
  url = {https://arxiv.org/abs/2607.14186}
}

@misc{shi2025taskcraft,
  author = {Shi, Dingfeng and Cao, Jingyi and Chen, Qianben and Sun, Weichen and Li, Weizhen and Lu, Hongxuan and Dong, Fangchen and Qin, Tianrui and Zhu, King and Liu, Minghao and Yang, Jian and Zhang, Ge and Liu, Jiaheng and Zhang, Changwang and Wang, Jun and Jiang, Yuchen Eleanor and Zhou, Wangchunshu},
  title = {{TaskCraft}: Automated Generation of Agentic Tasks},
  year = {2025},
  eprint = {2506.10055},
  archivePrefix = {arXiv},
  primaryClass = {cs.CL},
  doi = {10.48550/arXiv.2506.10055},
  url = {https://arxiv.org/abs/2506.10055}
}

@misc{xie2025agentsynth,
  author = {Xie, Jingxu and Xu, Dylan and Zhao, Xuandong and Song, Dawn},
  title = {{AgentSynth}: Scalable Task Generation for Generalist Computer-Use Agents},
  year = {2025},
  eprint = {2506.14205},
  archivePrefix = {arXiv},
  primaryClass = {cs.CL},
  doi = {10.48550/arXiv.2506.14205},
  url = {https://arxiv.org/abs/2506.14205}
}

@misc{ramrakhya2025autoplay,
  author = {Ramrakhya, Ram and Szot, Andrew and Attia, Omar and Yang, Yuhao and Nguyen, Anh and Mazoure, Bogdan and Gan, Zhe and Agrawal, Harsh and Toshev, Alexander},
  title = {Scaling Synthetic Task Generation for Agents via Exploration},
  year = {2025},
  eprint = {2509.25047},
  archivePrefix = {arXiv},
  primaryClass = {cs.AI},
  doi = {10.48550/arXiv.2509.25047},
  url = {https://arxiv.org/abs/2509.25047}
}

@misc{chen2026dive,
  author = {Chen, Aili and Zhang, Chi and Liu, Junteng and Chen, Jiangjie and Du, Chengyu and Li, Yunji and Zhong, Ming and Wang, Qin and Zhu, Zhengmao and Song, Jiayuan and Ji, Ke and He, Junxian and Zhao, Pengyu and Xiao, Yanghua},
  title = {{DIVE}: Scaling Diversity in Agentic Task Synthesis for Generalizable Tool Use},
  year = {2026},
  eprint = {2603.11076},
  archivePrefix = {arXiv},
  primaryClass = {cs.AI},
  doi = {10.48550/arXiv.2603.11076},
  url = {https://arxiv.org/abs/2603.11076}
}

@misc{bai2026clawgym,
  author = {Bai, Fei and Song, Huatong and Sun, Shuang and Cheng, Daixuan and Yang, Yike and Hao, Chuan and Li, Renyuan and Chang, Feng and Wei, Yuan and Tao, Ran and Dai, Bryan and Yang, Jian and Zhao, Wayne Xin and Wen, Ji-Rong},
  title = {{ClawGym}: A Scalable Framework for Building Effective Claw Agents},
  year = {2026},
  eprint = {2604.26904},
  archivePrefix = {arXiv},
  primaryClass = {cs.CL},
  doi = {10.48550/arXiv.2604.26904},
  url = {https://arxiv.org/abs/2604.26904}
}

@misc{pi2026data,
  author = {Pi, Renjie and Lam, Grace and Shoeybi, Mohammad and Jannaty, Pooya and Catanzaro, Bryan and Ping, Wei},
  title = {On Data Engineering for Scaling {LLM} Terminal Capabilities},
  year = {2026},
  eprint = {2602.21193},
  archivePrefix = {arXiv},
  primaryClass = {cs.CL},
  doi = {10.48550/arXiv.2602.21193},
  url = {https://arxiv.org/abs/2602.21193}
}

@misc{raoof2026openthoughts,
  author = {Raoof, Negin and Zhuang, Richard and Nezhurina, Marianna and Guha, Etash and Tejaswi, Atula and Marten, Ryan and Ruan, Charlie F. and Griggs, Tyler and Shaw, Alexander Glenn and Bansal, Hritik and Buchanan, E. Kelly and Gazizov, Artem and Heckel, Reinhard and Hegde, Chinmay and Jajee, Sankalp and Khazi, Daanish and Koukoumidis, Emmanouil and Li, Xiangyi and Liu, Hange and Natarajan, Shlok and Raj, Harsh and Roberts, Nicholas and Shen, Ethan and Singhi, Nishad and Siu, Michael and Suvarna, Ashima and Xing, Hanwen and Yubeaton, Patrick and Zhang, Robert and Chen, Leon Liangyu and Chen, Xiaokun and Dillmann, Steven and Gabriel, Saadia and Jiang, Xunyi and Kashyap, Anurag and Li, Boxuan and Park, Yein and Pham, Minh and Sanghavi, Sujay and Shi, Lin and Sun, Ke and Wang, Yixin and Xu, Zhiwei and Zhang, Erica and Zhao, Siyan and Zhao, Wanjia and Jitsev, Jenia and Dimakis, Alex and Feuer, Benjamin and Schmidt, Ludwig},
  title = {{OpenThoughts-Agent}: Data Recipes for Agentic Models},
  year = {2026},
  eprint = {2606.24855},
  archivePrefix = {arXiv},
  primaryClass = {cs.AI},
  doi = {10.48550/arXiv.2606.24855},
  url = {https://arxiv.org/abs/2606.24855}
}

@misc{shi2026facet,
  author = {Shi, Kou and Wang, Zun and Su, Qisheng and Huang, Shiting and Zhang, Ziao and Fang, Zhen and Ren, Qingnan and Liu, Jin and Zeng, Yu and Zhao, Yiming and Chen, Lin and Chen, Zehui and Zhao, Feng},
  title = {{FACET}: Preserving Source Intent and Executable State in Terminal Task Synthesis},
  year = {2026},
  eprint = {2608.18580},
  archivePrefix = {arXiv},
  primaryClass = {cs.AI},
  doi = {10.48550/arXiv.2608.18580},
  url = {https://arxiv.org/abs/2608.18580}
}

@misc{ding2026wildclawbench,
  author = {Ding, Shuangrui and Dai, Xuanlang and Xing, Long and Ding, Shengyuan and Liu, Ziyu and Yang, JingYi and Yang, Penghui and Zhang, Zhixiong and Wei, Xilin and Fang, Xinyu and Ma, Yubo and Duan, Haodong and Shao, Jing and Wang, Jiaqi and Lin, Dahua and Chen, Kai and Zang, Yuhang},
  title = {{WildClawBench}: A Benchmark for Real-World, Long-Horizon Agent Evaluation},
  year = {2026},
  eprint = {2605.10912},
  archivePrefix = {arXiv},
  primaryClass = {cs.CL},
  doi = {10.48550/arXiv.2605.10912}
}

@misc{ye2026claweval,
  author = {Ye, Bowen and Li, Rang and Yang, Qibin and Liu, Yuanxin and Yao, Linli and Lv, Hanglong and Xie, Zhihui and An, Chenxin and Li, Lei and Kong, Lingpeng and Liu, Qi and Sui, Zhifang and Yang, Tong},
  title = {{Claw-Eval}: Towards Trustworthy Evaluation of Autonomous Agents},
  year = {2026},
  eprint = {2604.06132},
  archivePrefix = {arXiv},
  primaryClass = {cs.AI},
  doi = {10.48550/arXiv.2604.06132}
}

@inproceedings{patwardhan2026gdpval,
  author = {Patwardhan, Tejal and Dias, Rachel and Proehl, Elizabeth and Kim, Grace and Wang, Michele and Watkins, Olivia and Fishman, Simon and Aljubeh, Marwan and Thacker, Phoebe and Fauconnet, Laurance and Kim, Natalie and Miserendino, Samuel and Chabot, Gildas and Li, David and Chao, Patrick and Sharman, Michael and Barr, Alexandra and Glaese, Amelia and Tworek, Jerry},
  title = {{GDPval}: Evaluating {AI} Model Performance on Real-World Economically Valuable Tasks},
  booktitle = {International Conference on Learning Representations},
  year = {2026},
  url = {https://proceedings.iclr.cc/paper_files/paper/2026/file/290c2430f91912204f30bbcc990fff1d-Paper-Conference.pdf}
}

@misc{vidgen2026apex,
  author = {Vidgen, Bertie and Mann, Austin and Fennelly, Abby and Stanly, John Wright and Rothman, Lucas and Burstein, Marco and Benchek, Julien and Ostrofsky, David and Ravichandran, Anirudh and Sur, Debnil and Venugopal, Neel and Hsia, Alannah and Robinson, Isaac and Huang, Calix and Varones, Olivia and Khan, Daniyal and Haines, Michael and Bridges, Austin and Boyle, Jesse and Twist, Koby and Richards, Zach and Mahapatra, Chirag and Foody, Brendan and Nitski, Osvald},
  title = {{APEX-Agents}},
  year = {2026},
  eprint = {2601.14242},
  archivePrefix = {arXiv},
  primaryClass = {cs.CL},
  doi = {10.48550/arXiv.2601.14242}
}

@article{sun2026agentslastexam,
  author = {Sun, Yiyou and Han, Xinyang and Zhang, Weichen and Pang, Yuanbo and Wang, Tianyu and Cao, Yuhan and Huang, Yixiao and Duroiu, Chris and Zhang, Haoyun and Lin, Jeffrey and Zhang, Weishu and Zeng, Tyler and Yan, Ying and Liu, Bo and Wen, Hanson and Xu, Mingyang and Liu, Xiaoyuan and Chen, Zimeng and Shi, Weiyan and Dsouza, Amanda and Chen, Vincent Sunn and Song, Dawn and Bryant, Patrick and Boettiger, Carl and Rangan, Yamini and Rothenberg, Bradley and Steinfeld, Kyle and Rao, Arvind and Schneider, Tapio and Yannakakis, Georgios and Zanna, Laure and Ozbay, Kaan and Sim, Ida and Zohdi, Tarek and Karniadakis, George Em and Gallant, Jack and Head-gordon, Teresa and others},
  title = {Agents' Last Exam},
  journal = {arXiv preprint arXiv:2606.05405},
  year = {2026},
  doi = {10.48550/arXiv.2606.05405}
}

@misc{tu2026scaleenv,
  author = {Tu, Dunwei and Hao, Hongyan and Yang, Hansi and Chen, Yihao and Zhang, Yi-Kai and Xia, Zhikang and Yang, Yu and Sun, Yueqing and Liu, Xingchen and Shen, Furao and Gu, Qi and Su, Hui and Cai, Xunliang},
  title = {{ScaleEnv}: Scaling Environment Synthesis from Scratch for Generalist Interactive Tool-Use Agent Training},
  year = {2026},
  eprint = {2602.06820},
  archivePrefix = {arXiv},
  primaryClass = {cs.AI},
  url = {https://arxiv.org/abs/2602.06820}
}

@misc{fan2026skillsynth,
  author = {Fan, Zhiyuan and Yu, Tinghao and Cai, Yuanjun and Guan, Jiangtao and Yang, Yun and Hu, Dingxin and Zhou, Jiang and Wu, Xing and Han, Zhuo and Zhang, Feng and Wang, Lilin},
  title = {Toward Scalable Terminal Task Synthesis via Skill Graphs},
  year = {2026},
  eprint = {2604.25727},
  archivePrefix = {arXiv},
  primaryClass = {cs.AI},
  url = {https://arxiv.org/abs/2604.25727}
}

@inproceedings{zhang2026agentv,
  author = {Zhang, Jiazheng and Fu, Ziche and Xi, Zhiheng and Jing, Wenqing and Chai, Mingxu and He, Wei and Zhang, Guoqiang and Fan, Chenghao and An, Chenxin and Chen, Wenxiang and Liu, Zhicheng and Pan, Haojie and Zhu, Dingwei and Gui, Tao and Zhang, Qi and Huang, Xuanjing},
  title = {{AgentV-RL}: Scaling Reward Modeling with Agentic Verifier},
  booktitle = {Findings of the Association for Computational Linguistics: ACL 2026},
  pages = {23078--23100},
  year = {2026},
  publisher = {Association for Computational Linguistics},
  doi = {10.18653/v1/2026.findings-acl.1156},
  url = {https://aclanthology.org/2026.findings-acl.1156/}
}

@inproceedings{Dou:2025:Sandbox,
  author       = {Shihan Dou and
                  Jiazheng Zhang and
                  Jianxiang Zang and
                  Yunbo Tao and
                  Weikang Zhou and
                  Haoxiang Jia and
                  Shichun Liu and
                  Yuming Yang and
                  Shenxi Wu and
                  Zhiheng Xi and
                  Muling Wu and
                  Rui Zheng and
                  Changze Lv and
                  Limao Xiong and
                  Shaoqing Zhang and
                  Lin Zhang and
                  Wenyu Zhan and
                  Rongxiang Weng and
                  Jingang Wang and
                  Xunliang Cai and
                  Yueming Wu and
                  Ming Wen and
                  Yixin Cao and
                  Tao Gui and
                  Xipeng Qiu and
                  Qi Zhang and
                  Xuanjing Huang},
  editor       = {Pushkar Mishra and
                  Smaranda Muresan and
                  Tao Yu},
  title        = {Multi-Programming Language Sandbox for LLMs},
  booktitle    = {Proceedings of the 63rd Annual Meeting of the Association for Computational
                  Linguistics (Volume 3: System Demonstrations), {ACL} 2025, Vienna,
                  Austria, July 27 - August 1, 2025},
  pages        = {40--50},
  publisher    = {Association for Computational Linguistics},
  year         = {2025},
  url          = {https://doi.org/10.18653/v1/2025.acl-demo.5},
  doi          = {10.18653/V1/2025.ACL-DEMO.5},
}

@article{zhang:2025:CRM,
  author       = {Jiazheng Zhang and
                  Wenqing Jing and
                  Zizhuo Zhang and
                  Zhiheng Xi and
                  Shihan Dou and
                  Rongxiang Weng and
                  Jiahuan Li and
                  Jingang Wang and
                  Mingxu Chai and
                  Shibo Hong and
                  Tao Gui and
                  Qi Zhang},
  title        = {Two Minds Better Than One: Collaborative Reward Modeling for {LLM}
                  Alignment},
  journal      = {CoRR},
  volume       = {abs/2505.10597},
  year         = {2025},
  url          = {https://doi.org/10.48550/arXiv.2505.10597},
  doi          = {10.48550/ARXIV.2505.10597},
  eprinttype   = {arXiv},
  eprint       = {2505.10597},
}

@inproceedings{xi2026agentgymrl,
  author = {Xi, Zhiheng and Huang, Jixuan and Liao, Chenyang and Huang, Baodai and Liu, Jiaqi and Guo, Honglin and Yang, Yajie and Zheng, Rui and Ye, Junjie and Zhang, Jiazheng and Chen, Wenxiang and He, Wei and Ding, Yiwen and Li, Guanyu and Chen, Zehui and Du, Zhengyin and Yao, Xuesong and Xu, Yufei and Chen, Jiecao and Gui, Tao and Wu, Zuxuan and Zhang, Qi and Huang, Xuanjing and Jiang, Yu-Gang},
  title = {{AgentGym-RL}: An Open-Source Framework to Train {LLM} Agents for Long-Horizon Decision Making via Multi-Turn {RL}},
  booktitle = {International Conference on Learning Representations},
  year = {2026},
  url = {https://openreview.net/forum?id=ZgCCDwcGwn}
}

@misc{zhang2026entropy,
  author = {Zhang, Jiazheng and Fu, Ziche and Shen, Junrui and Zhao, Yunbin and Zhang, Yunke and Xi, Zhiheng and Ma, Long and An, Chenxin and Zhang, Zhihao and Liu, Shichun and Zhu, Dingwei and Dou, Shihan and Liu, Shaofan and Li, Han and Zhou, Wiggin and Adams, Aiden and Gui, Tao and Huang, Fei and Zhang, Qi and Huang, Xuanjing},
  title = {Entropy Polarity in Reinforcement Fine-Tuning: Direction, Asymmetry, and Control},
  year = {2026},
  eprint = {2605.11775},
  archivePrefix = {arXiv},
  primaryClass = {cs.LG},
  url = {https://arxiv.org/abs/2605.11775}
}

@inproceedings{xi2026agentgym2,
  author = {Xi, Zhiheng and Yang, Dingwen and Liu, Jiaqi and Huang, Jixuan and Guo, Honglin and Huang, Baodai and Chen, Tinggang and Zhang, Qi and Lu, Zhonghang and Liu, Chenyu and Sun, Jiajun and Zhang, Jiazheng and Zhu, Dingwei and Guo, Xin and Wang, Junzhe and Zhang, Zhihao and Yang, Yuming and Ye, Junjie and Gao, Minghe and Liu, Dongrui and Ji, Jiaming and Li, Guohao and Gui, Tao and Zhang, Qi and Huang, Xuanjing},
  title = {{AgentGym2}: Benchmarking Large Language Model Agents in De-Idealized Real-World Environments},
  booktitle = {Proceedings of the 64th Annual Meeting of the Association for Computational Linguistics (Volume 1: Long Papers)},
  pages = {44451--44479},
  year = {2026},
  publisher = {Association for Computational Linguistics},
  doi = {10.18653/v1/2026.acl-long.2058},
  url = {https://aclanthology.org/2026.acl-long.2058/}
}

@misc{anthropic2025claudecode,
  author = {{Anthropic}},
  title = {Claude 3.7 Sonnet and Claude Code},
  year = {2025},
  month = feb,
  url = {https://www.anthropic.com/news/claude-3-7-sonnet}
}

@misc{openai2025codex,
  author = {{OpenAI}},
  title = {Introducing Codex},
  year = {2025},
  month = may,
  url = {https://openai.com/index/introducing-codex/}
}

@misc{qwen2026qwen35,
  author = {{Qwen Team}},
  title = {{Qwen3.5}: Towards Native Multimodal Agents},
  year = {2026},
  month = feb,
  url = {https://qwen.ai/blog?id=qwen3.5}
}

@misc{anthropic2026claudeopus5,
  author = {{Anthropic}},
  title = {Introducing {Claude Opus 5}},
  year = {2026},
  month = jul,
  url = {https://www.anthropic.com/news/claude-opus-5}
}

@misc{openai2026gpt56,
  author = {{OpenAI}},
  title = {{GPT-5.6}: Frontier Intelligence That Scales with Your Ambition},
  year = {2026},
  month = jul,
  url = {https://openai.com/index/gpt-5-6/}
}

@misc{doshi2026gemini37flash,
  author = {Doshi, Tulsee},
  title = {Introducing {Gemini 3.7 Flash}},
  year = {2026},
  month = aug,
  url = {https://blog.google/innovation-and-ai/models-and-research/gemini-models/introducing-gemini-3-7-flash}
}

@misc{deepseek2026v4,
  author = {{DeepSeek-AI} and others},
  title = {{DeepSeek-V4}: Towards Highly Efficient Million-Token Context Intelligence},
  year = {2026},
  eprint = {2606.19348},
  archivePrefix = {arXiv},
  primaryClass = {cs.CL},
  doi = {10.48550/arXiv.2606.19348},
  url = {https://arxiv.org/abs/2606.19348}
}

@misc{kimi2026k25,
  author = {{Kimi Team}},
  title = {{Kimi K2.5}: Visual Agentic Intelligence},
  year = {2026},
  eprint = {2602.02276},
  archivePrefix = {arXiv},
  primaryClass = {cs.CL},
  doi = {10.48550/arXiv.2602.02276},
  url = {https://arxiv.org/abs/2602.02276}
}

@misc{zai2026glm52,
  author = {{Z.ai}},
  title = {{GLM-5.2}: Built for Long-Horizon Tasks},
  year = {2026},
  month = jun,
  url = {https://z.ai/blog/glm-5.2}
}

@misc{minimax2026m3,
  author = {{MiniMax}},
  title = {{MiniMax M3}: Frontier Coding, {1M} Context, Native Multimodality---All in One Model},
  year = {2026},
  month = jun,
  url = {https://www.minimax.io/blog/minimax-m3}
}

@misc{steinberger2026openclaw,
  author = {Steinberger, Peter},
  title = {Introducing {OpenClaw}},
  year = {2026},
  month = jan,
  url = {https://openclaw.ai/blog/introducing-openclaw}
}

@misc{qwen2025qwen3coder,
  author = {{Qwen Team}},
  title = {{Qwen3-Coder}: Agentic Coding in the World},
  year = {2025},
  month = jul,
  url = {https://qwen.ai/blog?id=qwen3-coder}
}

@misc{minimax2026m27,
  author = {{MiniMax}},
  title = {{MiniMax M2.7}: Early Echoes of Self-Evolution},
  year = {2026},
  month = mar,
  url = {https://www.minimax.io/news/minimax-m27-en}
}
\bibliographystyle{iclr2027_conference}

\appendix
\section*{Appendix}
\section{Domain Taxonomy and Distribution}
\label{app:domain-taxonomy}

\subsection{Domain Taxonomy}
\begin{figure}[!htb]
  \centering
  \includegraphics[width=\linewidth]{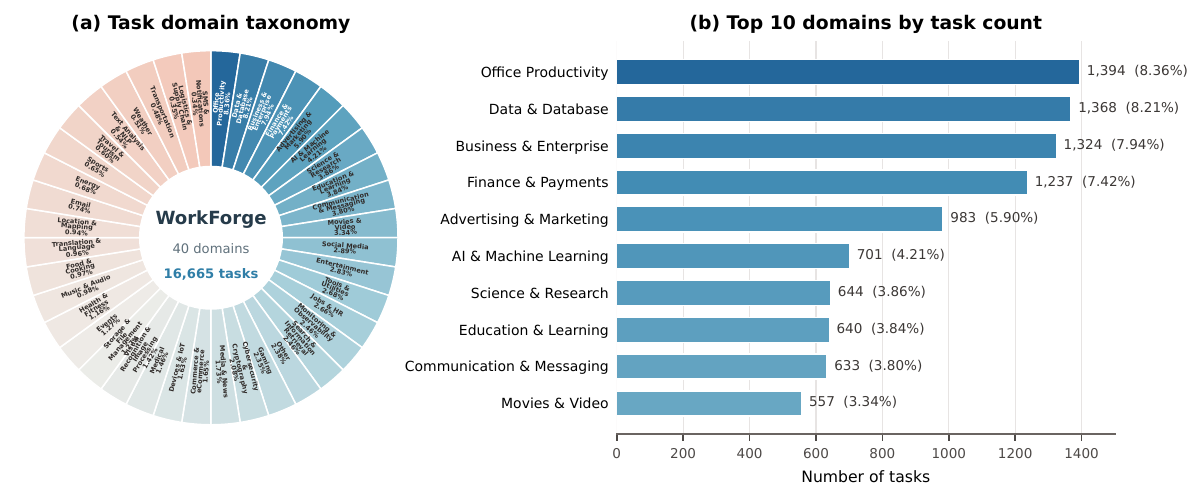}
  \caption{\textbf{Domain taxonomy and task distribution of the WorkForge SFT corpus.}
  The left panel shows all 40 domains, and the right panel reports the ten most represented domains.}
  \label{fig:domain-taxonomy-distribution}
\end{figure}

WorkForge organizes 16,665 SFT tasks into 40 domains, as shown in Figure~\ref{fig:domain-taxonomy-distribution}.
The taxonomy spans common workplace activities such as office productivity, data management, business operations, finance, and communication, while also covering software, scientific, media, healthcare, security, and infrastructure-oriented tasks.
The included workflows range from document and data processing to technical investigation and domain-specific decision making.

\subsection{Domain Distribution}
Table~\ref{tab:domain-taxonomy-statistics} lists the complete distribution.
The four most represented domains account for 31.9\% of the corpus, and the top ten account for 56.9\%.
The remaining 30 domains contribute 43.1\% of all tasks.
The distribution combines repeated coverage of broadly applicable workplace domains with coverage across specialized applications.

\begin{table}[!htb]
  \caption{\textbf{Complete domain distribution.} Counts are computed over 16,665 SFT tasks.}
  \label{tab:domain-taxonomy-statistics}
  \centering
  \scriptsize
  \setlength{\tabcolsep}{2.5pt}
  \renewcommand{\arraystretch}{1.08}
  \begin{tabularx}{\linewidth}{@{}rYrr@{\hspace{5pt}}rYrr@{}}
    \toprule
    \# & \textbf{Domain} & \textbf{Count} & \textbf{\%} &
    \# & \textbf{Domain} & \textbf{Count} & \textbf{\%} \\
    \midrule
    1 & Office Productivity & 1,394 & 8.36 & 21 & Commerce \& eCommerce & 275 & 1.65 \\
    2 & Data \& Database & 1,368 & 8.21 & 22 & Devices \& IoT & 271 & 1.63 \\
    3 & Business \& Enterprise & 1,324 & 7.94 & 23 & Medical & 243 & 1.46 \\
    4 & Finance \& Payments & 1,237 & 7.42 & 24 & Visual Recognition \& Image Processing & 237 & 1.42 \\
    5 & Advertising \& Marketing & 983 & 5.90 & 25 & Storage \& File Management & 212 & 1.27 \\
    6 & AI \& Machine Learning & 701 & 4.21 & 26 & Events & 211 & 1.27 \\
    7 & Science \& Research & 644 & 3.86 & 27 & Health \& Fitness & 194 & 1.16 \\
    8 & Education \& Learning & 640 & 3.84 & 28 & Music \& Audio & 163 & 0.98 \\
    9 & Communication \& Messaging & 633 & 3.80 & 29 & Food \& Cooking & 162 & 0.97 \\
    10 & Movies \& Video & 557 & 3.34 & 30 & Translation \& Language & 160 & 0.96 \\
    11 & Social Media & 482 & 2.89 & 31 & Location \& Mapping & 156 & 0.94 \\
    12 & Entertainment & 471 & 2.83 & 32 & Email & 123 & 0.74 \\
    13 & Tools \& Utilities & 447 & 2.68 & 33 & Energy & 113 & 0.68 \\
    14 & Jobs \& HR & 443 & 2.66 & 34 & Sports & 108 & 0.65 \\
    15 & Monitoring \& Observability & 410 & 2.46 & 35 & Travel \& Tourism & 100 & 0.60 \\
    16 & Search \& Information Retrieval & 410 & 2.46 & 36 & Text Analysis \& NLP & 90 & 0.54 \\
    17 & Other & 396 & 2.38 & 37 & Weather & 85 & 0.51 \\
    18 & Gaming & 391 & 2.35 & 38 & Transportation & 80 & 0.48 \\
    19 & Cybersecurity \& Cryptography & 346 & 2.08 & 39 & Logistics \& Supply Chain & 59 & 0.35 \\
    20 & Media \& News & 289 & 1.73 & 40 & SMS \& Notifications & 57 & 0.34 \\
    \bottomrule
  \end{tabularx}
\end{table}

\section{Pipeline Pseudocode and Prompt Templates}
\label{app:pipeline-prompts}

\subsection{Pipeline Pseudocode}
Algorithm~\ref{alg:workforge-pipeline} presents the end-to-end construction
procedure and follows the notation introduced in Section~\ref{sec:method}.

\begin{algorithm}[t]
\small
\SetKwInput{KwIn}{Input}
\SetKwInput{KwOut}{Output}
\SetKw{Return}{return}
\SetCommentSty{textbf}
\caption{\textbf{Authority-Separated Task Construction and Trajectory
Generation in \textsc{WorkForge}}}
\label{alg:workforge-pipeline}
\KwIn{Expert-authored professional workflow specifications; resource-acquisition
interface $\mathcal A_{\mathrm{ground}}$; solver policy $\pi_\theta$}
\KwOut{Grounded task packages $\mathcal T$; scored rollout corpus
$\mathcal D_{\mathrm{rollout}}$}
\linespread{1.06}\selectfont
$\mathcal T \leftarrow \varnothing;\quad
\mathcal D_{\mathrm{rollout}} \leftarrow \varnothing$;
\tcp{Phase 1: Workflow Specification}
Obtain expert-authored Scene Blueprints $\mathcal B=\{b_j\}_{j=1}^{M}$\;
\ForEach{Scene Blueprint $b_j \in \mathcal B$}{
    \tcp{Phase 2: Environment Synthesis}
    $\{R_{jk}\}_{k=1}^{K_j} \leftarrow
    \operatorname{GroundResourcePools}(b_j;\mathcal A_{\mathrm{ground}})$\;
    \ForEach{resource pool $R_{jk}$}{
        $W_{jk} \leftarrow \operatorname{Instantiate}(R_{jk})$\;
        $P_{jk} \leftarrow \operatorname{Explore}(W_{jk};b_j)$\;
        \tcp{Phase 3: Task Generation and Verification Design}
        Synthesize task instruction $q_{jk}$ and solution plan $h_{jk}$
        from $b_j$ and $P_{jk}$\;
        $\widehat F_{jk} \leftarrow
        \operatorname{VerifyFacts}(h_{jk};W_{jk},P_{jk})$\;
        $(\mathcal V_{jk}^{\mathrm{prog}},\mathcal V_{jk}^{\mathrm{sem}})
        \leftarrow \operatorname{BuildRubrics}
        (q_{jk},h_{jk},P_{jk},\widehat F_{jk})$\;
        $\mathcal V_{jk}\leftarrow
        \mathcal V_{jk}^{\mathrm{prog}}\cup\mathcal V_{jk}^{\mathrm{sem}}$\;
        Construct $G_{jk}^{\mathrm{prog}}$ from
        $\mathcal V_{jk}^{\mathrm{prog}}$ and $G_{jk}^{\mathrm{sem}}$ from
        $q_{jk}$ and $\mathcal V_{jk}^{\mathrm{sem}}$\;
        $\mathcal T_{jk}\leftarrow
        ((W_{jk},q_{jk}),(P_{jk},h_{jk},\widehat F_{jk},\mathcal V_{jk},
        G_{jk}^{\mathrm{prog}},G_{jk}^{\mathrm{sem}}))$\;
        $\mathcal T \leftarrow \mathcal T \cup \{\mathcal T_{jk}\}$\;
        \tcp{Phase 4: Trajectory Generation}
        $(\tau_{jk},Y_{jk})\leftarrow
        \operatorname{Rollout}(\pi_\theta;q_{jk},W_{jk})$\;
        $S_{jk}^{\mathrm{prog}}=\displaystyle
        \frac{1}{|\mathcal V_{jk}^{\mathrm{prog}}|}
        \sum_{v\in\mathcal V_{jk}^{\mathrm{prog}}}
        \mathbb I[v(W_{jk},Y_{jk})=1],\qquad
        S_{jk}^{\mathrm{sem}}=\displaystyle
        \frac{1}{|\mathcal V_{jk}^{\mathrm{sem}}|}
        \sum_{v\in\mathcal V_{jk}^{\mathrm{sem}}}\frac{s_v}{10}$\;
        $\mathcal D_{\mathrm{rollout}}\leftarrow
        \mathcal D_{\mathrm{rollout}}\cup
        \{(\tau_{jk},Y_{jk},S_{jk}^{\mathrm{prog}},
        S_{jk}^{\mathrm{sem}},\mu_{jk})\}$\;
    }
}
\Return $\mathcal T,\mathcal D_{\mathrm{rollout}}$\;
\end{algorithm}

\FloatBarrier
\section{Experimental, Training, and Evaluation Details}
\label{app:experimental-details}
\subsection{Supervised Fine-Tuning Setup}

We perform full-parameter supervised fine-tuning on 16,665 trajectories from
\textsc{WorkForge}-16k.
The training trajectories are generated by GLM 5.3 with the Claude Code
and OpenClaw harnesses.
We train Qwen3.5-35B-A3B-Base and the post-trained Qwen3.5-27B model for five
epochs using the same optimization recipe.
Table~\ref{tab:sft-hyperparameters} reports the training configuration.

\begin{table}[!htb]
  \caption{Supervised fine-tuning configuration used for the main models and
  training-data ablations.}
  \label{tab:sft-hyperparameters}
  \centering
  \small
  \setlength{\tabcolsep}{5pt}
  \renewcommand{\arraystretch}{1.12}
  \begin{tabular}{@{}ll@{}}
    \toprule
    \textbf{Configuration} & \textbf{Value} \\
    \midrule
    Optimizer & AdamW \\
    Adam coefficients & $\beta_1=0.9$, $\beta_2=0.95$ \\
    Peak learning rate & $1\times10^{-5}$ \\
    Weight decay & 0.1 \\
    Gradient clipping & 1.0 \\
    Learning-rate schedule & Cosine decay to $1\times10^{-7}$ \\
    Warmup ratio & 5\% of total steps \\
    Training epochs & 5 \\
    Global batch size & 64 sequences \\
    Micro-batch size & 1 sequence per GPU \\
    Maximum sequence length & 262,144 tokens \\
    Numerical precision & BF16 mixed precision \\
    Sequence packing & Enabled \\
    Dynamic batching & Enabled \\
    Gradient checkpointing & Enabled \\
    \bottomrule
  \end{tabular}
\end{table}

\subsection{Training Infrastructure and Data Formatting}

Training uses the \texttt{verl} supervised fine-tuning trainer with Ray~2.46,
PyTorch~2.7.1, CUDA~12.8, Transformers~5.10.4, and FlashAttention~2.8.1.
We use FSDP with Ulysses sequence parallelism of size four.  The
Qwen3.5-35B-A3B-Base runs use 64 NVIDIA H20 96GB GPUs, and the
Qwen3.5-27B runs use 128 GPUs.  Optimizer states are offloaded to CPU,
while model parameters remain on GPU.

Training examples are rendered incrementally from cumulative conversation
prefixes so that the reasoning content associated with each assistant turn is
preserved.  The loss is applied according to the per-message assistant loss
mask stored with each trajectory.  Variable-length examples are packed without
padding, with sequence boundaries retained through cumulative sequence-length
indices.  Dynamic batching limits each GPU batch by its token count, with a
maximum budget of 262,144 tokens.

\subsection{Inference Configuration}

We serve local models with vLLM and use the same decoding configuration for the \textsc{WorkForge} models and
the re-evaluated open-source baselines in Table~\ref{tab:main-results}.
Table~\ref{tab:inference-configuration} lists the shared settings.
The maximum context length is 262,144 tokens, whereas the 32,768-token limit
applies to each generated model response.  For models whose native context
window is shorter than 262,144 tokens, we use YaRN position extrapolation to
extend the context window to the shared limit.
Specifically, EnvScaler-Qwen3-8B, Arctic-AWM-14B, Nemotron-Terminal-32B,
and OpenThinkerAgent-32B use the Qwen3 architecture with the YaRN settings
in Table~\ref{tab:yarn-configuration}.

\begin{table}[!htb]
  \caption{Decoding configuration used for model evaluation.}
  \label{tab:inference-configuration}
  \centering
  \small
  \setlength{\tabcolsep}{7pt}
  \renewcommand{\arraystretch}{1.12}
  \begin{tabular}{@{}ll@{}}
    \toprule
    \textbf{Parameter} & \textbf{Value} \\
    \midrule
    Temperature & 0.7 \\
    Top-$p$ & 0.9 \\
    Top-$k$ & 20 \\
    Maximum generated tokens per response & 32,768 \\
    Maximum context length & 262,144 \\
    \bottomrule
  \end{tabular}
\end{table}

\begin{table}[!htbp]
  \caption{Shared YaRN configuration for the Qwen3-based baselines.}
  \label{tab:yarn-configuration}
  \centering
  \small
  \setlength{\tabcolsep}{7pt}
  \begin{tabular}{@{}ll@{}}
    \toprule
    \textbf{Parameter} & \textbf{Value} \\
    \midrule
    Maximum position embeddings & 262,144 \\
    RoPE base ($\theta$) & 1,000,000 \\
    RoPE scaling type & YaRN \\
    Scaling factor & 6.4 \\
    Original maximum position embeddings & 40,960 \\
    \bottomrule
  \end{tabular}
\end{table}

\subsection{Benchmark Protocols}
\label{app:benchmark}
All results in Table~\ref{tab:main-results} are obtained through our evaluation
runs on WildClawBench, Claw-Eval, GDPval, APEX-Agents, and Agents' Last Exam.
Table~\ref{tab:benchmark-protocols} summarizes their metrics and sampling protocols.
Pass@1 denotes the fraction of tasks passed in a single rollout under each benchmark's success criterion.
Claw-Eval uses three independent rollouts per task and reports Pass\symbol{94}3, the fraction of tasks passed in all three rollouts.
Unlike Pass@3, which requires at least one successful rollout, Pass\symbol{94}3 measures consistency across repeated runs.
All other benchmarks use one rollout per task.
WildClawBench and Claw-Eval use OpenClaw; GDPval and ALE use Claude Code;
APEX-Agents uses its official ReAct Toolbelt agent.
Table~\ref{tab:agent-evaluation-budgets} lists the harnesses, agent-level
wall-clock limits, and interaction budgets.

\begin{table}[!htb]
  \caption{Benchmark-specific metrics and sampling protocols.}
  \label{tab:benchmark-protocols}
  \centering
  \small
  \setlength{\tabcolsep}{4pt}
  \renewcommand{\arraystretch}{1.12}
  \begin{tabularx}{\linewidth}{@{}lYYc@{}}
    \toprule
    \textbf{Benchmark} & \textbf{Evaluation set} & \textbf{Reported metrics} &
    \textbf{Rollouts/task} \\
    \midrule
    WildClaw & Full set & Score, Pass@1 & 1 \\
    Claw-Eval & Full set & Score, Pass\symbol{94}3 & 3 \\
    GDPval & 201 text-only tasks & Rubric score & 1 \\
    APEX & Full set & Score, Pass@1 & 1 \\
    ALE & Full set & Score, Pass@1 & 1 \\
    \bottomrule
  \end{tabularx}
\end{table}

\begin{table}[!htbp]
  \caption{Evaluation harnesses and per-task agent execution budgets.}
  \label{tab:agent-evaluation-budgets}
  \centering
  \small
  \setlength{\tabcolsep}{4pt}
  \renewcommand{\arraystretch}{1.12}
  \begin{tabularx}{\linewidth}{@{}lYll@{}}
    \toprule
    \textbf{Benchmark} & \textbf{Harness} & \textbf{Wall-clock limit} &
    \textbf{Interaction limit} \\
    \midrule
    WildClaw & OpenClaw & 24,000\,s & Not set \\
    Claw-Eval & OpenClaw & 3,600\,s & 50 turns \\
    GDPval & Claude Code & 14,400\,s & 300 turns \\
    APEX & ReAct Toolbelt & 14,400\,s & 250 steps \\
    ALE & Claude Code & 28,800\,s & Unlimited \\
    \bottomrule
  \end{tabularx}
\end{table}

\subsection{GDPval Evaluation Subset}
\label{app:gdpval-subset}

We evaluate all models on the same 201-task subset of GDPval's official
220-task evaluation set, excluding 19 multimodal tasks.
Tasks are aligned using the evaluation payload's \texttt{case\_id}, which
corresponds to the official \texttt{task\_id}.
We report the official rubric score rather than the win-or-tie rate.
Table~\ref{tab:gdpval-excluded-tasks} lists the excluded tasks and their occupations.

\begin{table}[!htb]
  \caption{Multimodal tasks excluded from the GDPval evaluation set.
  Row numbers refer to the official 220-task set.}
  \label{tab:gdpval-excluded-tasks}
  \centering
  \footnotesize
  \setlength{\tabcolsep}{3pt}
  \renewcommand{\arraystretch}{1.12}
  \begin{tabularx}{\linewidth}{@{}rlY@{}}
    \toprule
    \textbf{Row} & \textbf{Task ID} & \textbf{Occupation} \\
    \midrule
    12 & \texttt{38889c3b-e3d4-49c8-816a-3cc8e5313aba} & Audio and Video Technicians \\
    13 & \texttt{ff85ee58-bc9f-4aa2-806d-87edeabb1b81} & Audio and Video Technicians \\
    14 & \texttt{4b894ae3-1f23-4560-b13d-07ed1132074e} & Audio and Video Technicians \\
    56 & \texttt{e222075d-5d62-4757-ae3c-e34b0846583b} & Film and Video Editors \\
    57 & \texttt{c94452e4-39cd-4846-b73a-ab75933d1ad7} & Film and Video Editors \\
    58 & \texttt{75401f7c-396d-406d-b08e-938874ad1045} & Film and Video Editors \\
    59 & \texttt{a941b6d8-4289-4500-b45a-f8e4fc94a724} & Film and Video Editors \\
    79 & \texttt{4d1a8410-e9c5-4be5-ab43-cc55563c594c} & First-Line Supervisors of Office \& Admin Support \\
    102 & \texttt{b9665ca1-4da4-4ff9-86f2-40b9a8683048} & Industrial Engineers \\
    111 & \texttt{46fc494e-a24f-45ce-b099-851d5c181fd4} & Mechanical Engineers \\
    127 & \texttt{3baa0009-5a60-4ae8-ae99-4955cb328ff3} & News Analysts, Reporters, and Journalists \\
    134 & \texttt{e6429658-4de1-42dd-a9e0-2d2b9b02fb10} & Nurse Practitioners \\
    147 & \texttt{57b2cdf2-ad62-4591-aa91-aad489740320} & Private Detectives and Investigators \\
    149 & \texttt{a46d5cd2-55fe-48fa-a4c6-6aaf6b9991b5} & Private Detectives and Investigators \\
    153 & \texttt{e4f664ea-0e5c-4e4e-a0d3-a87a33da947a} & Producers and Directors \\
    171 & \texttt{11593a50-734d-4449-b5b4-f8986a133fd8} & Real Estate Sales Agents \\
    172 & \texttt{94925f49-36bc-42da-b45b-61078d329300} & Real Estate Sales Agents \\
    188 & \texttt{f2986c1f-2bbf-4b83-bc93-624a9d617f45} & Pharmacists \\
    203 & \texttt{fe0d3941-e32c-4bf1-a643-b566d2b4cb3c} & Sales Representatives, Wholesale \& Manufacturing \\
    \bottomrule
  \end{tabularx}
\end{table}

\subsection{GDPval Rubric-Level Analysis}
\label{app:gdpval-rubric-analysis}

We group GDPval criteria into five deliverable-quality categories and compute the
pass rate within each category. For category $c$ and model $m$, let
$\mathcal{I}_c$ denote its criteria and $z_j^{(m)}\in\{0,1\}$ indicate whether
criterion $j$ is satisfied. The category pass rate is
\begin{equation}
  R_c^{(m)} =
  \frac{\sum_{j\in\mathcal{I}_c} z_j^{(m)}}
       {|\mathcal{I}_c|}.
\end{equation}
We report $100R_c^{(m)}$ as a percentage and compute each gain as
$100(R_c^{(\mathrm{SFT})}-R_c^{(\mathrm{Base})})$ percentage points.
Table~\ref{tab:gdpval-rubric-categories} reports the number and share of criteria
assigned to each category together with the resulting pass rates.

\begin{table}[!htb]
  \caption{GDPval rubric pass rates across five deliverable-quality categories.}
  \label{tab:gdpval-rubric-categories}
  \centering
  \footnotesize
  \setlength{\tabcolsep}{3.5pt}
  \renewcommand{\arraystretch}{1.10}
  \begin{tabularx}{\linewidth}{@{}Xrrrrr@{}}
    \toprule
    \textbf{Category} & \textbf{Criteria} & \textbf{Share} & \textbf{Base} &
    \textbf{SFT} & \textbf{$\Delta$} \\
    \midrule
    Visual / layout quality & 371 & 4.2\% & 74.1\% & 87.1\% & +13 pp \\
    Numerical claims & 523 & 5.9\% & 64.6\% & 72.8\% & +8.2 pp \\
    Cross-source reconciliation & 103 & 1.2\% & 86.4\% & 92.2\% & +5.8 pp \\
    Reasoning / argumentation & 200 & 2.2\% & 82.0\% & 86.5\% & +4.5 pp \\
    Content coverage & 7,718 & 86.6\% & 77.4\% & 83.4\% & +6.0 pp \\
    \bottomrule
  \end{tabularx}
\end{table}

\FloatBarrier
\section{Additional Ablation Analyses}
\label{app:additional-ablations}

The training-data ablations use Qwen3.5-35B-A3B-Base with the supervised
fine-tuning recipe in Table~\ref{tab:sft-hyperparameters}.

\subsection{Data Volume Scaling}
\label{app:data-scaling}

We train models at four data scales: 1K, 3.6K, 8K, and 16K trajectories.
For each target scale, we independently sample the requested number of
trajectories from the full corpus; subsets at different scales are not nested.
The 16K setting uses all 16,665 trajectories.
Figure~\ref{fig:scaling-analysis}(a) reports the results on ALE and WildClaw.

\subsection{Trajectory-Length Scaling}
\label{app:turn-scaling}

We construct the turn-scaling subsets from trajectories generated with the
Claude Code harness.
Trajectories are grouped into three disjoint intervals by their number of turns,
and we randomly sample 1,000 trajectories from each interval
(Table~\ref{tab:turn-scaling-subsets}).
Figure~\ref{fig:scaling-analysis}(b) reports performance on ALE and WildClaw,
while Table~\ref{tab:turn-scaling-behavior} summarizes tool-use behavior on WildClaw.

\begin{table}[!htb]
  \caption{Training subsets for trajectory-length scaling.}
  \label{tab:turn-scaling-subsets}
  \centering
  \small
  \setlength{\tabcolsep}{12pt}
  \begin{tabular}{@{}lrr@{}}
    \toprule
    \textbf{Group} & \textbf{Turns per trajectory} & \textbf{Trajectories} \\
    \midrule
    G1 & 0--20 & 1,000 \\
    G2 & 21--40 & 1,000 \\
    G3 & 41--60 & 1,000 \\
    \bottomrule
  \end{tabular}
\end{table}

\begin{table}[!htb]
  \caption{Tool-use behavior under trajectory-length scaling on WildClaw. Longer-turn supervision increases exploration intensity, breadth, and productive execution.}
  \label{tab:turn-scaling-behavior}
  \centering
  \footnotesize
  \setlength{\tabcolsep}{5pt}
  \renewcommand{\arraystretch}{1.08}
  \begin{tabular}{@{}lccccc@{}}
    \toprule
    \textbf{Group} & \shortstack{\textbf{Productive}\\\textbf{calls}}
    & \shortstack{\textbf{Skill}\\\textbf{discovery}}
    & \shortstack{\textbf{Tool}\\\textbf{calls}}
    & \shortstack{\textbf{Distinct}\\\textbf{actions}}
    & \shortstack{\textbf{Tool types}\\\textbf{used}} \\
    \midrule
    G1 & 11.7 & 21.0\% & 17.7 & 17.0 & 3.3 \\
    G2 & 23.1 & 21.9\% & 38.3 & 33.0 & 3.6 \\
    G3 & \textbf{24.0} & \textbf{33.3\%} & \textbf{42.0} & \textbf{38.3} & \textbf{4.0} \\
    \bottomrule
  \end{tabular}
\end{table}

\subsection{Verification-Filter Ablation}
\label{app:verification-filter}

We construct three 1,000-trajectory training sets from the same pool of completed
rollouts using the filtering rules in Table~\ref{tab:verification-filter-rules}.
The LLM-only variant selects trajectories by the semantic score
$S^{\mathrm{sem}}$, the programmatic-only variant uses $S^{\mathrm{prog}}$,
and the hybrid variant uses their equally weighted mean.
Each variant applies a threshold of 0.7 to its selection score.
The downstream results on ALE and APEX are reported in Table~\ref{tab:verification-ablation}.

\begin{table}[!htb]
  \caption{Trajectory selection rules for the verification-filter ablation.}
  \label{tab:verification-filter-rules}
  \centering
  \small
  \setlength{\tabcolsep}{7pt}
  \renewcommand{\arraystretch}{1.12}
  \begin{tabular}{@{}llr@{}}
    \toprule
    \textbf{Variant} & \textbf{Selection rule} & \textbf{Trajectories} \\
    \midrule
    LLM only & $S^{\mathrm{sem}} \geq 0.7$ & 1,000 \\
    Programmatic only & $S^{\mathrm{prog}} \geq 0.7$ & 1,000 \\
    Hybrid & $0.5S^{\mathrm{sem}} + 0.5S^{\mathrm{prog}} \geq 0.7$ & 1,000 \\
    \bottomrule
  \end{tabular}
\end{table}

\FloatBarrier
\section{Data Examples}
\label{app:data-examples}

\subsection{Cross-Region Factor-Premium Analysis}
\label{app:factor-analysis-example}

This Finance \& Payments task asks a quantitative strategist to assess factor
premia and prepare an investment-committee briefing.
WorkForge constructs the workspace from the public Kaggle dataset
\texttt{nikitamanaenkov/famafrench-factors-and-portfolios}, a local mirror of
the Kenneth French data library with an August 2025 vintage.
The example connects workspace exploration, evidence-grounded verification,
cross-region analysis, and the production of three linked deliverables.

\subsubsection{Task Instruction}

\begin{promptbox}{Task Instruction}
You are a quantitative strategist at a multi-asset manager. The investment committee
is weighing a new factor-tilt allocation and wants evidence on whether the classic
equity factors still earn premia -- and where (US vs developed vs emerging markets).
Using the firm's local mirror of the Kenneth French data library, build the analysis
and a decision-ready briefing.

Assess each major factor's long-run and recent performance, compare premia across
regions over a fair common window, and recommend which factors and regions deserve
capital (and which to avoid). Be explicit about data caveats that affect your
conclusions.

Deliver:
- results/factor_briefing.pptx -- committee briefing with findings and a clear
  recommendation
- results/factor_analysis.xlsx -- supporting performance tables
- results/factor_returns_tidy.csv -- the cleaned return series behind your analysis

Explore the workspace to determine structure, coverage, and pitfalls yourself.
\end{promptbox}

\subsubsection{Workspace and Source Materials}

The workspace contains 297 CSV files in 16 directories, with a maximum nesting
depth of six.
It includes factor series, sorted portfolios, research breakpoints, and industry
returns, organized into US, Developed, Emerging, Industry, and Other branches.
Momentum data also appear outside the US factor directory.

\begin{promptbox}{Workspace Layout}
workspace/
|-- data/
|   |-- US/
|   |   |-- Factors/                         [5 files]
|   |   |-- Portfolios/
|   |   |   |-- Bivariate/                   [49 files]
|   |   |   `-- Univariate/                  [18 files]
|   |   `-- Research_Breakpoints_Data/       [8 files]
|   |-- Developed/
|   |   |-- Factors/                         [36 files]
|   |   `-- Portfolios/                      [114 files]
|   |-- Emerging/
|   |   |-- Factors/                         [2 files]
|   |   `-- Portfolios/                      [7 files]
|   |-- Industry/                           [24 files]
|   `-- Other/                              [34 files]
`-- results/                                [initially empty]
\end{promptbox}

Individual files contain descriptive preambles and multiple frequency blocks.
For example, the US three-factor file stacks monthly observations and annual
returns in the same CSV:

\begin{promptbox}{Source Excerpt: F-F\_Research\_Data\_Factors.csv}
This file was created using the 202508 CRSP database.
The 1-month TBill rate data until 202405 are from Ibbotson Associates.
Starting from 202406, the 1-month TBill rate is from ICE BofA US 1-Month Treasury Bill Index.

,Mkt-RF,SMB,HML,RF
192607,   2.89,  -2.55,  -2.39,   0.22
192608,   2.64,  -1.14,   3.81,   0.25
...
202508,   1.85,   3.87,   4.41,   0.38

 Annual Factors: January-December
1927,  29.47,  -2.46,  -3.75,   3.12
1931, -44.84,   3.07, -13.69,   1.07
1933,  56.49,  44.61,  31.28,   0.30
\end{promptbox}

Other files declare missing-value sentinels of $-99.99$ or $-999$.
The US files identify CRSP as their source, whereas Developed and Emerging files
identify Bloomberg.
The solver must separate frequency blocks, handle missing values, reconcile
factor names such as \texttt{Mom} and \texttt{WML}, and determine overlapping
valid observations across regions.

\subsubsection{Workspace Evidence and Verification Criteria}

Workspace inspection produces 12 usable fact records with computation references.
One record corrects the assumption that international series share a July 1990
start: the Developed series begins in July 1990, but the Emerging five-factor
file begins in July 1989.
The final factual criterion references this corrected record.

\begin{promptbox}{Selected Workspace Fact Records}
vf01: US three-factor monthly series starts at 192607.
      First row: 192607, 2.89, -2.55, -2.39, 0.22.
vf03: The US three-factor CSV contains a monthly block followed by an annual block.
vf07: File headers declare missing values as -99.99 or -999.
vf09: US files use CRSP; Developed and Emerging files use Bloomberg.
vf12 (corrected): Developed starts at 199007; Emerging five-factor data start at 198907.
      Computation: inspect the first data rows of the two regional factor files.
\end{promptbox}

The task has 23 criteria: eight programmatic checks and 15 semantic criteria.
Programmatic checks assess deliverable validity and specific reported facts.
Semantic criteria assess analysis, data handling, evidence-to-recommendation
consistency, and deliverable quality.
Selected criteria are shown below.

\begin{promptbox}{Selected Verification Criteria}
DC-2 | Programmatic | file_valid
The analysis workbook is valid and contains at least one populated sheet.

FA-4 | Programmatic | value_in_context | fact_ref: vf12
The Emerging market 5-factor series start (198907) is correctly distinguished from
the Developed start of 199007.
Expected value: 198907; context keywords: Emerging, 1989, start.

AD-2 | Semantic
Compare regional premia over a common window, rather than each region's full history.

HC-1 | Semantic
Separate monthly and annual blocks so annual returns do not enter monthly statistics.

HC-2 | Semantic
Treat -99.99 and -999 as missing values rather than numerical returns.

HC-3 | Semantic
Discuss the CRSP/Bloomberg source difference when interpreting regional comparisons.

DQ-3 | Semantic
Produce a tidy long-format CSV with date, region, factor, and return fields.
\end{promptbox}

\subsubsection{Trajectory Excerpts}

GLM 5.3 executes the task with OpenClaw in 1,317 seconds, producing
121 messages and 60 tool calls.
Early exploration checks both ends of the source files and then compares regional
schemas, provenance, and date coverage.

\begin{promptbox}{Rollout Excerpts: Inspection and Data Handling}
[9] Assistant
Now let me inspect the actual file formats -- French CSVs are notoriously
messy (preamble text, multiple stacked tables, annual sections, footers,
-99.99 missing codes).

[15] Assistant
Key pitfalls already visible:
- US is CRSP (1926+), Developed/Emerging are Bloomberg (1989/1990+)
- Emerging has `-99.99` missing RMW/CMA early
- whitespace-padded values
- momentum is a separate file (`Mom` for US, `WML` for intl)
- stacked monthly+annual tables
Let me check the toolchain and the Emerging RMW/CMA validity start.
\end{promptbox}

The solver checks factor-level availability next.
Emerging RMW becomes available in July 1991 and CMA in July 1992, making
July 1992 the binding start for the cross-region five-factor panel.

\begin{promptbox}{Rollout Excerpt: Common-Window Construction}
[32] Assistant
Clean panel built. The binding constraint for a fair all-factor,
all-region window is Emerging CMA starting 1992-07, so the common
window is 1992-07 to 2025-08 (398 months). Now the analytics engine.

Tool call: write /project/scripts/analytics.py
Compute factor performance analytics from results/factor_returns_tidy.csv.
Outputs (results/tables/*.csv) consumed by the xlsx/pptx builders, plus
results/_headline.json with key numbers for the briefing.
\end{promptbox}

\end{document}